\documentclass[lettersize,journal]{IEEEtran}
\usepackage[markup=default]{changes}
\definechangesauthor[name={ZKC}, color=orange]{z}
\definechangesauthor[name={JH}, color=blue]{jh}
\definechangesauthor[name={JY}, color=red]{jy}

\usepackage{amsmath,amsfonts,amssymb,bm}
\usepackage{algorithmic}
\usepackage{algorithm}
\usepackage{array}
\usepackage{textcomp}
\usepackage{stfloats}
\usepackage{url}
\usepackage{verbatim}
\usepackage{color}
\usepackage{subcaption}
\usepackage{graphicx}
\usepackage{cite}
\usepackage{xcolor}
\usepackage{tabularx} 
\usepackage{ragged2e} 
\usepackage{booktabs} 
\usepackage{multirow}
\newtheorem{defn}{\textbf{Definition}}

\DeclareMathAlphabet\mathbfcal{OMS}{cmsy}{b}{n}

\usepackage{etoolbox}
\renewcommand{\footnoterule}{
    \kern -3pt
    \hrule width 0.2\textwidth height 0.4pt
    \kern 2pt
}
\makeatletter
\patchcmd{\@makecaption}
  {\scshape}
  {}
  {}
  {}

\makeatother
\usepackage[justification=justified]{caption}
\begin{document}

\title{Multi-Granular Attention based Heterogeneous Hypergraph Neural Network}

\author{Hong Jin, Kaicheng Zhou, Jie Yin, Lan You, Zhifeng Zhou

\thanks{This work was supported in part by the National Natural Science Foundation of China (No.62377009) and the Technology Innovation Special Program of Hubei Province (No.2024BAB034). \textit{(Corresponding author: Lan You.)}}
\thanks{Kaicheng Zhou and Zhifeng Zhou are with the School of Computer Science, Hubei University, Wuhan 430062, China, 18772668057@163.com }
\thanks{Hong Jin is with Hubei Key Laboratory of Big Data Intelligent Analysis and Application (Hubei University), Wuhan 430062, China , anya@hubu.edu.cn}
\thanks{Jie Yin is with the Discipline of Business Analytics, The University of Sydney, Sydney, NSW 2006, Australia, jie.yin@sydney.edu.au}
\thanks{Lan You is with Key Laboratory of Intelligent Sensing System and Security (Hubei University), Ministry of Education, Wuhan 430062, China, yoyo@hubu.edu.cn}
}
\maketitle



\begin{abstract}
Heterogeneous graph neural networks (HeteGNNs) have demonstrated strong abilities to learn node representations by effectively extracting complex structural and semantic information in heterogeneous graphs. 
Most of the prevailing HeteGNNs follow the neighborhood aggregation paradigm, leveraging meta-path based message passing to learn latent node representations. 
However, due to the pairwise nature of meta-paths, these models fail to capture high-order relations among nodes, resulting in suboptimal performance. 
Additionally, the challenge of ``over-squashing'', where long-range message passing in HeteGNNs leads to severe information distortion, further limits the efficacy of these models. 
To address these limitations, this paper proposes MGA-HHN, a Multi-Granular Attention based Heterogeneous Hypergraph Neural Network for heterogeneous graph representation learning. MGA-HHN introduces two key innovations: (1) a novel approach for constructing meta-path based heterogeneous hypergraphs that explicitly models higher-order semantic information in heterogeneous graphs through multiple views, and (2) a multi-granular attention mechanism that operates at both the node and hyperedge levels. This mechanism enables the model to capture fine-grained interactions among nodes sharing the same semantic context within a hyperedge type, while preserving the diversity of semantics across different hyperedge types. As such, MGA-HHN effectively mitigates long-range message distortion and generates more expressive node representations.
Extensive experiments on real-world benchmark datasets demonstrate that MGA-HHN outperforms state-of-the-art models, showcasing its effectiveness in node classification, node clustering and visualization tasks. 

\end{abstract}


\section{Introduction}
\IEEEPARstart
Heterogeneous graphs, also known as heterogeneous information networks (HINs), consist of multiple types of nodes and relations \cite{r25,r32}.
Graphs in real-world scenarios, such as social networks\cite{r32}, bibliographic networks\cite{gtn} and World Wide Web~\cite{hu2020heterogeneous,li2023hypergraph}, are typically heterogeneous and contain diverse semantic information. 
Representation learning in heterogeneous graphs aims to learn a mapping function that projects the input nodes into a lower-dimensional space while preserving the heterogeneous structure and rich semantics inherent in heterogeneous graphs~\cite{r7,r10,r27,metapath2vec,metagraph2vec}. 
How to learn informative node representations of heterogeneous graphs is a key research problem to facilitate various downstream graph analysis tasks, such as node classification, node clustering or graph classification, with successful applications across recommendation systems~\cite{r11}, drug prediction, healthcare systems~\cite{r12}, and beyond.

Graph Neural Networks (GNNs) \cite{r9} have demonstrated strong performance in graph representation learning and gained widespread attention in recent years. Representative GNNs like Graph Convolutional Networks (GCNs)~\cite{gcn} and Graph Attention Networks (GATs)~\cite{gat} are primarily designed for homogeneous graphs, which contain only one type of nodes and edges. These models, however, cannot be directly applied to heterogeneous graphss, which consist of multiple types of nodes and edges. 
In such heterogeneious graphs, different node and edge types possess distinct attributes residing in different feature spaces. Moreover, the topological structure in heterogeneous graphs often carries rich semantics, and a node's local neighborhood can vary significantly depending on the type of relations involved. This inherent heterogeneity poses significant challenges for effective representation learning in heterogeneous graphs.

\begin{figure*}[htbp]
\centering
\includegraphics[width=\textwidth]{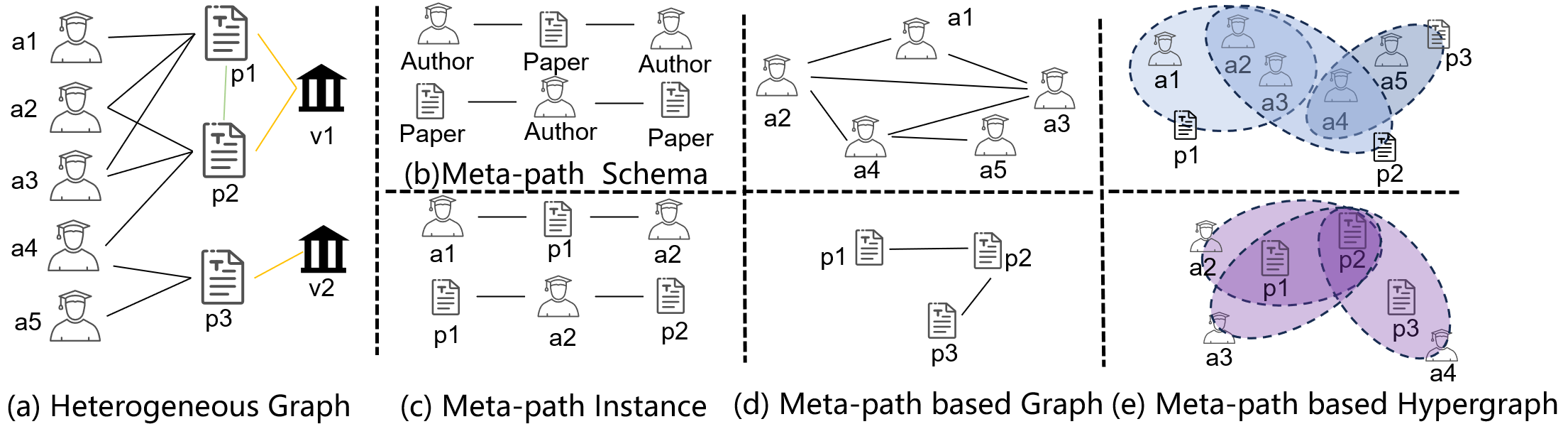}
\caption{An illustration of a heterogeneous graph, meta-paths and hypergraphs. (a) A heterogeneous graph with three types of nodes: Authors ($A$), Papers ($P$), and Venues ($V$). (b) The Author-Paper-Author ($APA$) and Paper-Author-Paper ($PAP$) meta-path schemas. (c) Instances of the Author-Paper-Author ($APA$) and Paper-Author-Paper ($PAP$) meta-paths. (d) Conventional graphs based on the meta-paths $APA$ and $PAP$. (e) Hypergraph based on the meta-paths $APA$ and $PAP$, where each hyperedge (marked by a dashed circle) captures a multivariate relationship among nodes (e.g., authors a1, a2 and a3 coauthor the same paper p1). In contrast, a meta-path based conventional graph decomposes these multivariate relationships into binary and pairwise relations (e.g., Author a1 coauthors with a2, a1 with a3, and a2 with a3).}
\label{fg:illustration}
\end{figure*}

To address the inherent heterogeneity, Heterogeneous Graph Neural Networks (HeteGNNs) have been proposed to learn node representations in heterogeneous graphs.
Most existing HeteGNNs exploit message-passing schemes to accommodate various types of nodes and edges, learning individual node representations by aggregating information from meta-path based neighbors. Although meta-path based methods have achieved state-of-the-art performance in many downstream tasks, they often fall short in capturing the complex multivariate relations inherently present in heterogeneous graphs. Specifically, a meta-path, defined as an ordered sequence of node types and edge types, can only represent pairwise relationships between nodes~\cite{pathsim}. In graph theory, one edge in a conventional graph simply represents a pairwise connection between two nodes. However, 
interactions between nodes in real-world heterogeneous graphs are not limited to pairwise relations, but could extend to ternary, quaternary, or even higher-order connections. For example, in a bibliographic network shown in Fig.~\ref{fg:illustration}, collaboration between two authors on a single paper can be modeled as an edge in a meta-path that reflects a pairwise coauthorship. 
However, in practice, multiple authors commonly collaborate on a single paper, forming multivariate relationships that extend beyond simple pairwise structures. Many scenarios in bibliographic networks feature such non-pairwise relationships \cite{r19}, such as ``multiple papers published in the same venue" or hierarchical complex high-order relations like ``multiple authors publishing papers in the same venue". These intricate relations are beyond the representation capacity of current meta-path based methods. Therefore, it is essential to develop an advanced representation learning paradigm that can effectively capture multivariate relationships in heterogeneous graphs.

Furthermore, most HeteGNNs rely on the message-passing paradigm, using learnable nonlinear functions to propagate information across a heterogeneous graph~\cite{rgcn,HAN,magnn,hegnn,gtn,r18}. However, this paradigm suffers from several limitations, particularly the issue of over-squashing, where messages from distant nodes become increasingly distorted during propagation. As highlighted in recent studies \cite{osq,hegnn,HAN,magnn,rgcn}, Message Passing Neural Networks (MPNNs) often struggle with graph learning tasks involving long-range dependencies, especially in graphs with many distant nodes in long-range neighborhoods. This difficulty arises because the representations of many distant nodes must be compressed into fixed-size vectors during neighborhood propagation, resulting in the over-squashing phenomenon. For most HeteGNNs, meta-path based neighbors are used to guide the selection of nodes for information aggregation when generating the target node's final representation. As the network depth increases, the size of the receptive field grows exponentially, introducing long-range dependencies. Consequently, as messages propagate across the network, information from non-adjacent nodes becomes increasingly distorted, degrading the quality of the learned representations.

To address the aforementioned issues, we propose a novel framework, Multi-Granular Attention based Heterogeneous Hypergraph Neural Network (MGA-HHN), for heterogeneous graph representation learning. MGA-HHN effectively captures the complex semantics and high-order relations inherent in heterogeneous graphs to learn informative node representations. The MGA-HHN framework comprises three key components.
(1) \textbf{Meta-path based Heterogeneous Hypergraph Construction}: MGA-HHN introduces a novel approach for constructing meta-path based heterogeneous hypergraphs that explicitly models rich, higher-order semantics in heterogeneous graphs through multiple views.
Specifically, each symmetric meta-path is conceptualized as a distinct type of hyperedge to capture semantic relationships, where the central node type of a symmetric meta-path defines the type of semantic hyperedge. For instance, node instances sampled along a symmetric meta-path $APA$ form a hyperedge of type $P$, while nodes along $APVPA$ form a hyperedge of type $V$. These different types of semantic hyperedges are naturally combined to form a unified heterogeneous hypergraph, as illustrated in Fig.~\ref{fg2}(b).
(2) \textbf{Node-level attention}: To enhance representation learning, MGA-HHN decomposes a heterogeneous hypergraph into multiple views, each corresponding to a specific hyperedge type that captures a distinct high-order semantic. Within each view, MGA-HHN employs a transformer-based self-attention mechanism to model long-range node-node interactions, rather than merely focusing on proximate nodes in local neighborhoods. Structural information is further incorporated via the view-specific hypergraph adjacency matrix. This node-level attention mechanism effectively alleviates the risk of information distortion from distant nodes during message passing, yielding more expressive and semantically meaningful node representations. (3) \textbf{Hyperedge-level attention}: To integrate node representations across different views, MGA-HHN introduces a hyperedge-level attention mechanism that adaptively assesses the relative importance of different hyperedge types for a given target node. This allows the model to adaptively weighs and aggregates node representations learned from various semantics perspectives, ensuring that the final node representations capture the full spectrum of semantic relationships in heterogeneous graphs. By integrating both node-level and hyperedge-level attention, the multi-granular attention mechnisam in MGA-HHN enables to effectively learn node representations that encapsulate rich, fine-grained semantics in heterogeneous graphs.


The main contributions of this paper are summarized as follows:
\begin{itemize}
\item We introduce an effective method for constructing meta-path based heterogeneous hypergraphs, enabling to explicitly model high-order semantic information in the original heterogeneous graph. 

\item We propose a novel multi-granular attention based heterogeneous hypergraph neural network (MGA-HHN) for heterogeneous graph representation learning. By integrating a multi-granular mechanism at both the node and hyperedge levels, MGA-HHN can more effectively handle long-range dependencies and complex relational structures within heterogeneous graphs, overcoming the limitations of existing HeteGNNs.

\item We conduct extensive experiments on three real-world datasets to validate the effectiveness of MGA-HHN. Comparisons with seven representative baselines, including homogeneous and heterogeneous GNNs, demonstrate the superiority of our proposed method.

\end{itemize}

\section{Related works}
This section reviews two key research areas related to our work, including heterogeneous graph neural networks and hypergraph neural networks.

\subsection{Heterogeneous Graph Neural Networks}
To extend deep neural networks to representation learning in heterogeneous graphs, researchers have proposed a series of Heterogeneous Graph Neural Networks (HeteGNNs)~\cite{r7,r10,r26}. Most existing HeteGNNs are primarily based on meta-paths and can be categorized into two groups: predefined meta-path based methods and automatically learned meta-path based methods. Predefined meta-path based methods rely on domain expertise to design specific meta-paths that reflect semantic relationships in heterogeneous graphs and then learn node representations by aggregating information from meta-path based neighbors. Wang et al.~\cite{HAN} proposed a Heterogeneous graph Attention Network (HAN), which employs node-level and semantic-level attention to capture the importance of nodes and meta-paths for information aggregation. The node-level attention measures the importance between a node and its meta-path based neighbors, while semantic-level attention learns the relative importance of different meta-paths. Inspired by HAN, Fu et al.~\cite{magnn} proposed a meta-path Aggregation Graph Neural Network (MAGNN), which introduces intra-meta-path and inter-meta-path  aggregation mechanisms. Unlike node-level aggregation in HAN, MAGNN's intra-meta-path aggregation incorporates the semantics of intermediate nodes along a meta-path, and inter-meta-path aggregation combines messages from multiple meta-paths.

While meta-path based models have achieved superior performance in heterogeneous graph representation learning, they rely on predefined meta-paths which requires domain expertise. To alleviate the need for presetting meta-paths, Yun et al.~\cite{gtn} proposed Graph Transformer Networks (GTN), which automatically extracts useful meta-paths in heterogeneous graphs. GTN partitions the heterogeneous graph into several subgraphs according to relation types and feeds subgraph adjacency matrices into a multi-channel GCN to learn node representations. 
Hong et al.~\cite{r34} proposed a Heterogeneous Graph Structural Attention Neural Network (HetSANN) to directly encode structural information of a heterogeneous graph. HetSANN designs a type-aware attention layer that aggregates neighbors’ features with different relation importance. Similarly, Hu et al.~\cite{hu2020heterogeneous} proposed a Heterogeneous Graph Transformer (HGT), which utilizes a meta-relation triplet attention mechanism to parameterize the weight matrices for calculating the attention score over each edge. This mechanism allows HGT to automatically and implicitly learn and extract meta-paths. Although these models achieve promising results, they have primarily focused on capturing binary and pairwise relations, struggling to fully leverage high-order relations inherently present in heterogeneous graphs. 

\subsection{Hypergraph Neural Networks}

A hypergraph generalizes the conventional graph structure by allowing a hyperedge to connect multiple nodes simultaneously, making it well-suited for capturing complex high-order relations~\cite{r31,r46}. Recently, several hypergraph neural networks have been proposed to leverage this capacity for graph representation learning. For example, Feng et al.~\cite{hgnn} proposed a Hypergraph Neural Network (HGNN), which defines a spectral convolution operator in a hypergraph, facilitating efficient information propagation between nodes by exploiting high-order relations and local clustering structures. Yadati et al.~\cite{hypergcn} proposed HyperGCN, which approximates each hyperedge as a set of pairwise edges and updates node representations using a GCN model. Bai et al.~\cite{bai} combined hypergraph convolution with an attention mechanism to enhance representation learning. However, these models are primarily designed for homogeneous hypergraphs, where all nodes and hyperedges belong to the same type, and thus they cannot be directly applied to capture diverse semantic relationships in heterogeneous graphs.

Several recent studies have made steps forward to explore node representation learning for heterogeneous hypergraphs.
Tu et al.~\cite{r33} proposed Deep Hyper-Network Embedding (DHNE) to learn node representations in heterogeneous hypergraphs by defining first-order and second-order node proximities. These promximities measure the $N$-tuplewise similarity between nodes, preserving rich structural information. However, this method is limited to uniform hypergraphs.
Ding et al.~\cite{r43} applied an attention mechanism to capture heterogeneous high-order context information of words on text hypergraph, while Sun et al.~\cite{hwnn} introduced wavelet basis functions into hypergraph convolution to improve computational efficiency. However, these methods are constrained by using a single weight matrix for different node and hyperedge types.
Fan et al.~\cite{hvae} proposed a Heterogeneous Hypergraph Variational Autoencoder (HeteHG-VAE), which first maps a heterogeneous graph to a heterogeneous hypergraph and then by using variational inference to learn deep latent representations of nodes and hyperedges. Yet, HeteHG-VAE defines hyperedges based on events, which cannot extract high-order relations with composite semantics, such as papers coauthored by multiple authors and published in the same venue. Li et al.~\cite{li2023hypergraph} introduced a Hypergraph Transformer Neural Network (HGTN), which constructs a heterogeneous hypergraph from the original meta-paths and uses an attention mechanism to weight these meta-paths, enabling the automatic generation of new meta-paths. HGTN then employs a multi-scale attention module on the hypergraph convolution to learn higher-order node representations. However, it constructs heterogeneous hypergraph by considering only the two end nodes in a meta-path, which inevitably results in information loss. 

In summary, these models often rely on stacking multiple layers to capture high-order relations in heterogeneous graphs. However, as model depth increases, they often suffer from over-squashing, where information from distant nodes becomes excessively compressed into fixed-size representations. This leads to a significant loss of contextual and semantic richness, particularly in  heterogeneous hypergraphs where complex, multi-relational structures are common. To address this challenge, there is a pressing need for more advanced GNNs that can effectively model high-order relations while mitigating the over-squashing phenomenon. Our work is thus proposed to fill this critical gap, aiming to learn more expressive and semantically meaningful node representations in heterogeneous graphs. 


\section{Preliminaries and problem statement}
In this section, we first introduce necessary preliminaries and definitions, followed by a formal definition of heterogeneous graph representation learning.  

\subsection{Preliminaries and Definitions}

\begin{defn}
\textbf{Heterogeneous Graph}. A \textit{heterogeneous graph} is defined as a directed graph $\mathcal{G} = \left \{ \mathcal{V},\mathcal{E}, \mathcal{O},\mathcal{R}, \phi, \varphi \right\}$ with multiple types of nodes or edges, where $\mathcal{V}$ denotes the set of nodes and $\mathcal{E}$ denotes the set of edges. 
Each node $v$ is associated with a $d$-dimensional feature vector $\mathbf{x}\in \mathbb{R}^d$. The node feature matrix is denoted as $\mathbf{X} \in \mathbb{R}^{|\mathcal{V}| \times d}$.
A heterogeneous graph is also associated with a node type mapping function $\phi: \mathcal{V} \rightarrow  \mathcal{O}$ and an edge type mapping function $\varphi: \mathcal{E}\rightarrow \mathcal{R}$, where $\mathcal{O}$ and $\mathcal{R}$ denote the sets of predefined node types and edge types, respectively, where $\left | \mathcal{O} \right |+ \left | \mathcal{R} \right |> 2$. 
\end{defn}

Fig.~\ref{fg:illustration}(a) shows an example bibliographic heterogeneous graph, which consists of several types of nodes: Author ($A$), Paper ($P$), Venue ($V$), and relations (co-authorship relation between authors and publishing relation between papers and venues).
 
In heterogeneous graphs, two nodes can be connected via different semantic paths, which are called meta-paths.

\begin{defn} 
\textbf{Meta-path}. A \textit{meta-path}  $\varPsi$ is defined as a path in the form of $ O_{1} \xrightarrow{R_{1}}O_{2}\xrightarrow{R_2}\cdots \xrightarrow{R_{l}}O_{\mathit{l+1}}$, (abbreviated as $O_{1} O_{2} \cdots O_{{l+1}}$), 
which describes a composite relation $R = R_1 \circ R_2 \circ \cdots \circ R_l$ between node types $O_1$  and $O_{l+1}$ where $\circ$ denotes the composition operator on relations.
\end{defn} 

As illustrated in Fig.~\ref{fg:illustration}(a), two authors can be connected via multiple meta-paths, such as Author-Paper-Author ($APA$) and Author-Paper-Venue-Paper-Author ($APVPA$). Each meta-path typically captures distinctive semantics. For example, $APA$ reflects a co-authorship relation, while $APVPA$ indicates that papers written by different authors are published in the same venue.

A meta-path instance $\psi $ is a specific node sequence in a heterogeneous graph that instantiates the meta-path schema defined by $\varPsi$.

\begin{defn}
\textbf{Heterogeneous Hypergraph}. A \textit{heterogeneous hypergraph} can be represented as $\mathcal{G}_h = \left \{ \mathcal{V},\mathcal{E},\mathcal{T}_v,\mathcal{T}_e, f_{\mathcal{V}}, f_{\mathcal{E}} \right \}$, where $\mathcal{V}$ denotes a set of nodes with a node type mapping function $f_{\mathcal{V}}: \mathcal{V} \rightarrow  \mathcal{T}_v$, and $\mathcal{E}$ denotes a set of hyperedges with a hyperedge type mapping function $f_{\mathcal{E}}: \mathcal{E}\rightarrow \mathcal{T}_e$. $\mathcal{T}_v$ and $\mathcal{T}_e$ denote the set of node types and hyperedge types, such as $|\mathcal{T}_v| + |\mathcal{T}_e| > 2$. For any hyperedge $e \in \mathcal{E}$, it can be denoted as $\left \{ v_1,v_2,\cdots,v_k\right \} \subseteq \mathcal{V}$.   
The relationship between nodes and hyperedges can be indicated by an incidence matrix $\mathbf{H}\in \mathbb{R}^{|\mathcal{V}|\times |\mathcal{E}|}$, where $\mathbf{H}(v,e) = 1$ if hyperedge $e$ is incident with node $v$, and $\mathbf{H}(v,e) = 0$ otherwise. 
\end{defn} 

In our work, we leverage the expressive power of heterogeneous hypergraphs by proposing an effective meta-path based construction method that explicitly captures higher-order semantic information in heterogeneous graphs through multiple views (Section~\ref{subsectioin:construction}). This approach enables the modeling of high-order relations among multiple nodes and preserving multi-type semantics inherent in heterogeneous graphs.


\subsection{Problem Statement}

Given a heterogeneous graph $\mathcal{G} = \left \{ \mathcal{V}, \mathcal{E} ,
\mathcal{O}, \mathcal{R}, \phi, \varphi \right \}$, the objective of learning heterogeneous network representation learning is to learn a mapping function $\Omega : \mathcal{V} \rightarrow \mathbb{R}^{\widetilde{d}}$ that embeds each node $v \in \mathcal{V}$ into a low-dimensional space $\mathbb{R}^{\widetilde{d}}$ while preserving both network structure and semantic information in $\mathcal{G}$. Nodes with similar structure and semantics in $\mathcal{G}$ are expected to have representations that are close to each other in the low-dimensional space.

The symbols and notations commonly used in this paper are summarized in Table~\ref{tab1}.

\begin{table}[h]
\caption{Main Notations and Descriptions}
\label{tab1}
\begin{tabular}{ll}
\toprule
Notations                                                                                                                                                                                                                                    & Description                                    \\ \midrule
$ \mathcal{G}  $                                                                 & a heterogeneous graph                              \\ 
$\varPsi$                                                                                    & Meta-path                                         \\ 
K                                                                                    &  Number of task-related meta-paths.                \\ 
$ \mathcal{G}_h $                                                                    & Heterogeneous hypergraph                         \\ 
$\mathbf{H} $                                                       & The hypergraph incidence matrix                  \\ 
$\mathcal{V},\mathcal{E}$                                                      &The sets of nodes/edges               \\ 
$\mathcal{O},\mathcal{R}$                                    &The sets of node/edge types                             \\
$O,R$                                                        &Certain node/edge type   \\
$\mathbf{X}$                                                    & $\mathbf{X}$ is the node feature matrix\\
$v,\mathbf{x}  $                                                 & A node and its feature vector                                    \\ 
 $\widetilde{d} $                                                       &  The dimension of node representations                 \\ 
$\mathbf{W} $                                                              & Learnable weight Matrix                     \\
$\bm{\alpha}^r$                                                           & The node-level attention matrix in a 
single view  \\
${\beta}^r$                                                       & Hyperedge-level attention score in a 
single view\\
$\mathbf{Z}$                                                      & The final node representations \\

\bottomrule
\end{tabular}
\end{table}


\begin{figure*}[htbp]
\centering
\includegraphics[width=0.8\textwidth]{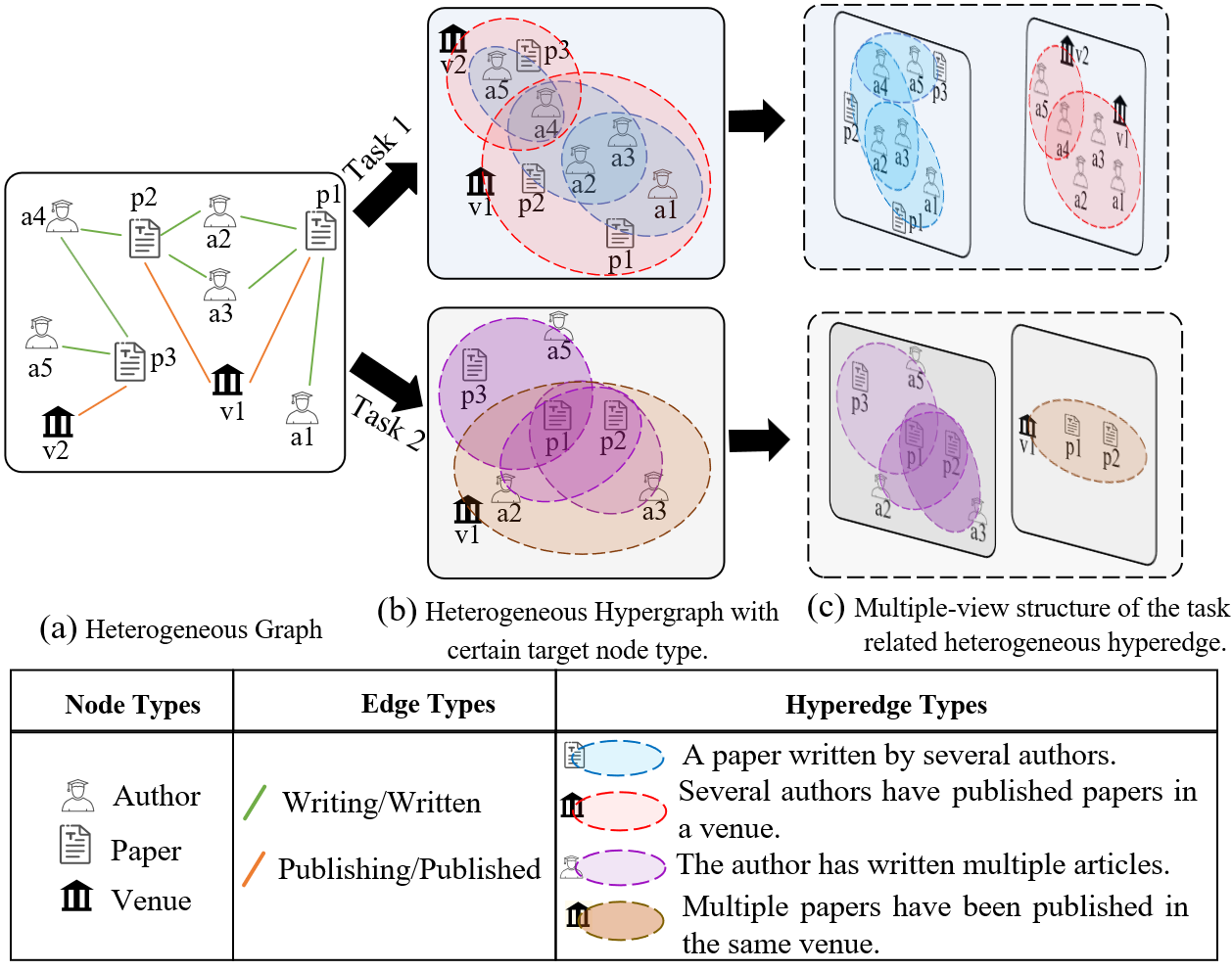}
\caption{Customized heterogeneous hypergraph  for specific downstream tasks. (a) An example heterogeneous graph with three types of nodes: Authors ($A$), Papers ($P$) and Venues ($V$). (b) Customized heterogeneous hypergraphs based on task-specific symmetric meta-paths. (c) Multi-view representation of the corresponding task-related  heterogeneous hypergraph. } 
\label{fg2}
\end{figure*}

\section{The Proposed Method}


In this section,  we first introduce the overall framework of MGA-HHN. After that, we give a detailed description of key components in the framework. 
\begin{figure*}[t]
\centering
\includegraphics[width=\textwidth]{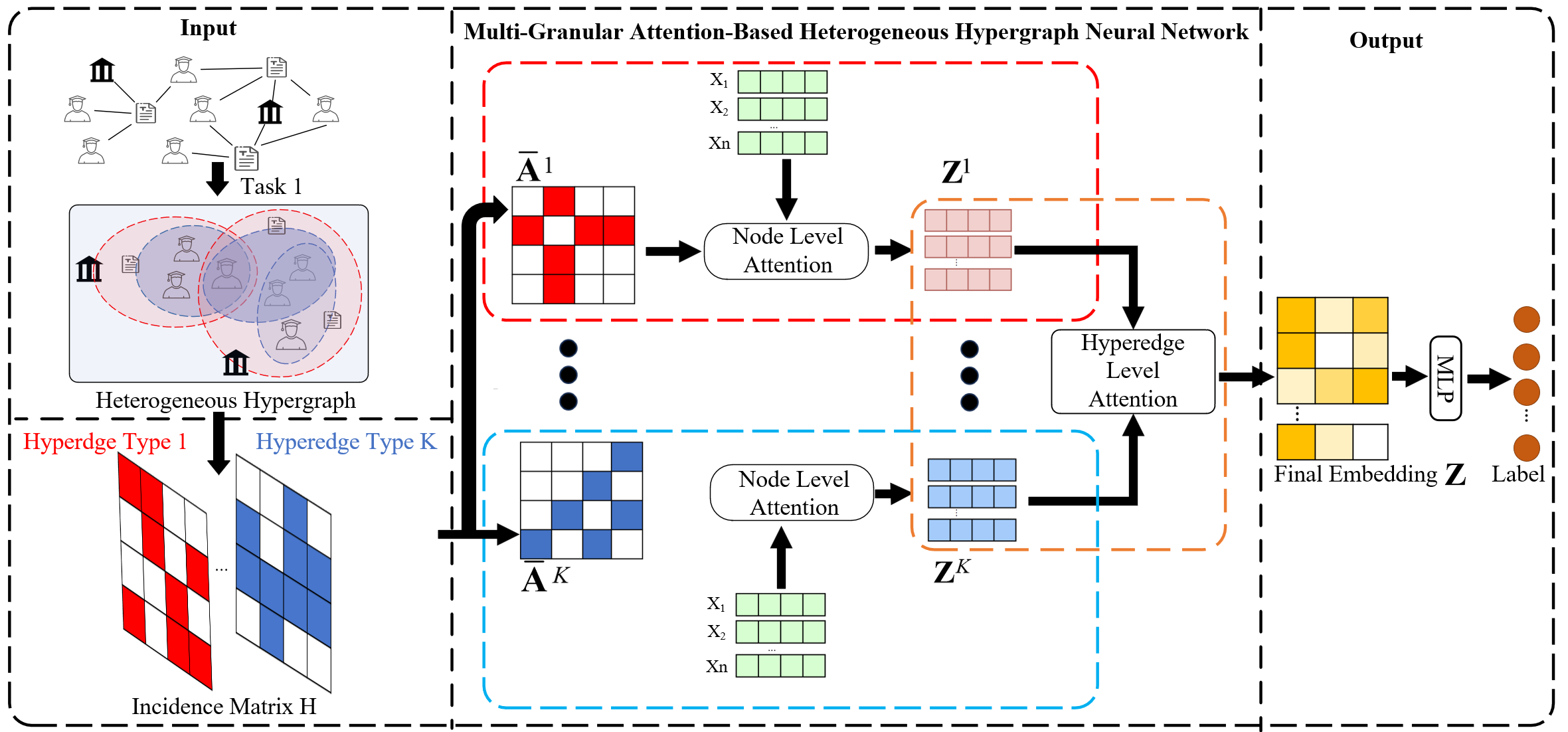}
\caption{The overall framework of MGA-HHN. First, a heterogeneous hypergraph is constructed based on the task-related meta-paths with  certain target node type determined by a downstream task. Next, for each hyperedge type, node-level attention is employed to learn node representations within a single view. Finally, hyperedge-level attention is used to fuse the representations of the target nodes from multiple views.}
\label{fg3}
\end{figure*}

\subsection{Main Idea and Overall Framework}

Fig.~\ref{fg3} shows the overall workflow of the proposed MGA-HHN framework for learning node representations in a heterogeneous graph. The main idea of MGA-HHN is to explicitly extract high-order semantic information from the original heterogeneous graph, modeled as a multi-view heterogeneous hypergraph, and introduces a multi-granular attention mechanisam for more effective node representation learning. In a nutshell,
MGA-HHN comprises three key components: (1) Heterogeneous hypergraph construction, which constructs a multi-view heterogeneous hypergraph from the original heterogeneous graph based on task-related symmetric meta-paths (Definition~\ref{def:symmetric_metapath}). (2) Node-level attention, which learns node representations within a single view defined by one distinct type of hyperedge. This enables each node to attend to all other nodes in a hypergraph, capturing important node-node interactions beyond local neighborhoods. (3) Hyperedge-level attention, which fuses $K$ different node representations learned from multiple views $\mathbf{Z} = \left \{ \mathbf{Z}^{1},\mathbf{Z}^{2},\cdots ,\mathbf{Z}^{K}\right \}$ to induce the final node representations. 
 
In what follows, we present the details of the three key components, respectively.

\subsection{Heterogeneous Hypergraph Construction}
\label{subsectioin:construction}

Consider a heterogeneous graph $\mathcal{G} = \left \{ \mathcal{V}, \mathcal{E},\mathcal{O}, \mathcal{R}, \phi, \varphi \right \} $, for any target node type $O \in \mathcal{O}$, determined by a specific downstream task, a heterogeneous hypergraph can be constructed using task-specific meta-paths. Below, we first introduce the core definitions necessary for our construction process.

\begin{defn} 
\label{def:symmetric_metapath}
\textbf{Symmetric Meta-path}. A meta-path $\varPsi $ is \textit{symmetric} if its corresponding composite relation $R$ is symmetric.  
\end{defn}

In heterogeneous graphs, many meta-paths are symmetric. For example, the meta-path $APA$ (Author–Paper–Author) is symmetric, while the meta-path $VPA$ (Venue–Paper–Author) is asymmetric. As shown in Fig.~\ref{fg:illustration}, authors a1 and a2 co-author the same paper p1, and thus a1--p1--a2 forms  a symmetric meta-path instance of $APA$. 

\begin{defn}
\textbf{Identifier Node and Slave Node}. In a symmetric meta-path, the central node type uniquely determines the type of a hyperedge, which is also referred to as the \textit{identifier node}. A \textit{slave node} is any member node of a hyperedge that, together with the identifier node, represents the complete semantics of the hyperedge. 
\end{defn}

The two end slave nodes of a symmetric meta-path, are considered as the target nodes. A slave node is not necessarily unique to a single hyperedge and may be shared across multiple hyperedges. For example, for meta-path $APA$, a hypergraph can be constructed using $P$ as the hyperedge and $A$ as the target node. For meta-path $APVPA$, a hypergraph can be constructed using $V$ as the hyperedge, $A$ as the target node, and $P$ as the slave node.

Given a symmetric meta-path, we treat the identifier node type as the hyperedge type, where nodes share the same semantic context.  To construct the heterogeneous hypergraph, we first sample all instances of the corresponding meta-paths. Then, for each instance of the identifier node type, we respectively group its connected end nodes to form
a hyperedge. This construction process enables to model high-order, multivariate relationships among target nodes while preserving the semantics encoded in different meta-paths.

The resulting heterogeneous hypergraph can be formalized as $\mathcal{G}_h = \left \{ \mathcal{V}_h, \mathcal{E}_h,\mathcal{T}_v,\mathcal{T}_e, f_{\mathcal{V}}, f_{\mathcal{E}} \right \}$, where $\mathcal{V}_h$ represents a set of nodes associated with the target node type $O \in \mathcal{O}$. 
This heterogeneous hypergraph can be decomposed as
$\mathcal{G}_h= \bigcup_{r \in \mathcal{T}_e}\mathcal{G}^r_h$, where each $\mathcal{G}^r_h$ is a heterogeneous hypergraph defined by a specific hyperedge type $r \in \mathcal{T}_e$.
As a result, the collection $\left \{ \mathcal{G}^r_h|r \in \mathcal{T}_e\right \}$, with $\left | \mathcal{T}_e \right | = K $, forms a multi-view network representation, where $\mathcal{G}^r_h$ is a hypergraph with a homogeneous hyperedge type in each view. The customized heterogeneous hypergraph can be represented by a set of incidence matrices $\left \{ \mathbf{H}^{r} \right \}_{r=1}^{K}$. Each matrix $\mathbf{H}^{r}$ represents the incidence relation under the $r$-th hyperedge type. 

Fig.~\ref{fg2}(b) shows an illustrative example of a customized heterogeneous hypergraph constructed from the original heterogeneous graph in Fig.~\ref{fg2}(a). The heterogeneous hypergraph comprise nodes of the target node type, $\mathcal{T}_v= \{\textrm{Author}\} $, and two types of hyperedges, $\mathcal{T}_e=\left \{ \textrm{Paper}, \textrm{Venue} \right \}$, defined by the identifier node P and V of meta-path $APA$ and $APVPA$, respectively. Fig.~\ref{fg2}(c) illustrates the multi-view representation of the corresponding heterogeneous hypergraph.

For $\mathcal{G}^r_h$ within a single view, its incidence matrix $\mathbf{H}^r$ and adjacency matrix $\mathbf{A}^r$ can be defined as follows:
\begin{align}
\mathbf{H}^r(v,e) &= \begin{cases} 1, & \textrm{if} \ v\in e, \\ 
0,& \textrm{otherwise},
\end{cases}\\
\mathbf{D}_{e}^r &=\displaystyle\sum_{v\in e}\mathbf{H}^r\left ( v,e\right ),\\
\mathbf{A}^r &= \left ({\mathbf{H}}^r \right )^{\mathrm{T}}\mathbf{W}^r_{e}({\mathbf{D}^r_{e}})^{-1}{\mathbf{H}^r},
\end{align}
where $ \mathbf{W}^r_{e}$  and $\mathbf{D}^r_{e}$ denote the diagonal weight and degree matrix of hyperedges in the incidence matrix $\mathbf{H}^r$, respectively. Let $\mathbf{\bar{A}}^r$ denote the normalization of $\mathbf{A}^r$ as:
\begin{equation}
\begin{aligned}
\mathbf{\bar{A}}^r &= (\mathbf{D}^r_v)^{-\frac{1}{2}}\left (\mathbf{H}^r \right )^{\mathrm{T}}\mathbf{W}^r_{e}(\mathbf{D}^r_{e})^{-1}{\mathbf{H}^r} {(\mathbf{D}^r_{v})}^{\tfrac{1}{2}},
\end{aligned}
\end{equation}
where $ \mathbf{D}^r_{v} \in \mathbb{R}^{|\mathcal{V}|\times |\mathcal{V}|}$ denotes the diagonal matirx of node degrees:
\begin{equation}
\begin{aligned}
\mathbf{D}^r_{v}=\displaystyle\sum_{e\in {\mathcal{E}^r}}\mathbf{W}^r_{e}\mathbf{H}^r\left ( v,e\right).
\end{aligned}
\end{equation}

For each hyperedge type $r \in \mathcal{T}_e$, we can obtain the normalized adjacency matrix $\mathbf{\bar{A}}^r$ within a single view.

\subsection{Node-level Attention}

\begin{figure*}[htbp]
\centering
\includegraphics[ width=\textwidth]{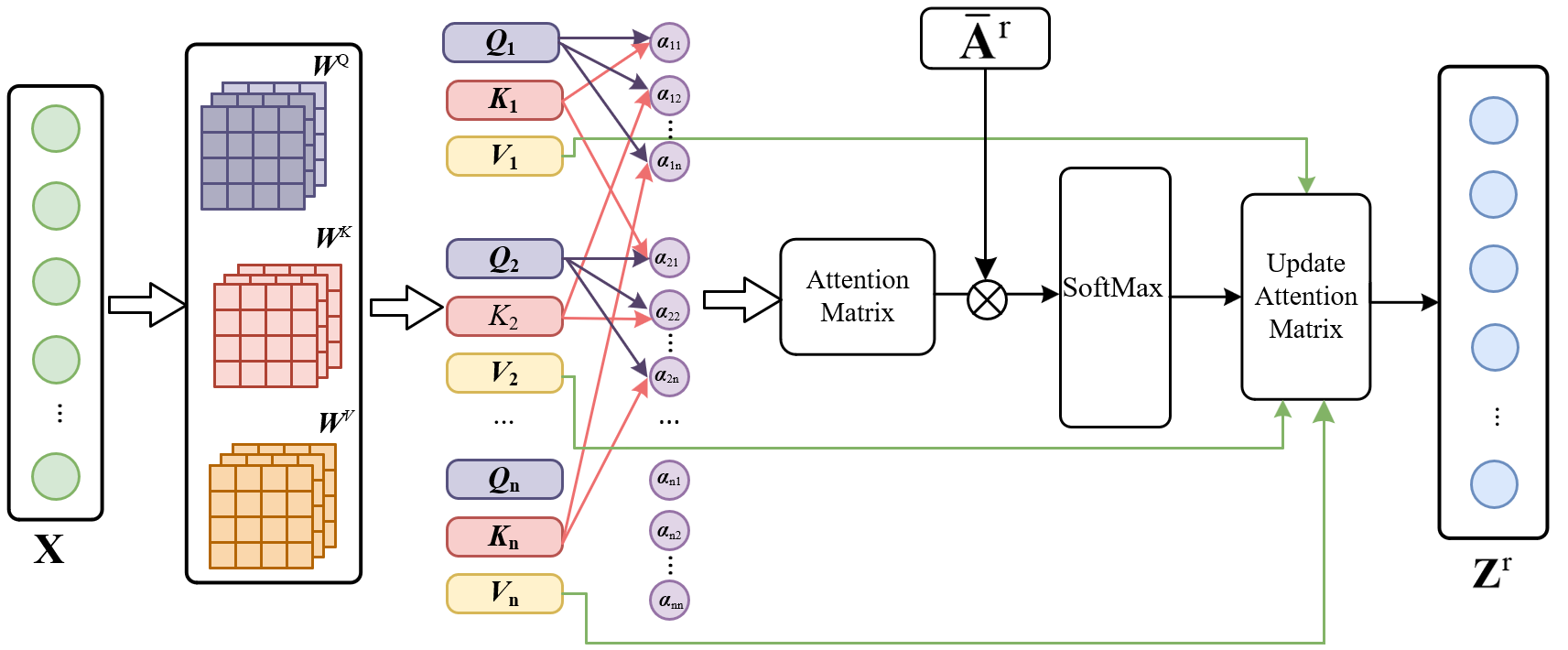}
\caption{Node-level attention. First, we use $ \mathbf{W}^{Q}, \mathbf{W}^{K}, \mathbf{W}^{V} $ to project node features into query   $\left ( \mathbf{Q} \right ) $,  key  $\left ( \mathbf{K}\right )$ and value $\left ( \mathbf{V}\right )$ matrices. Then we use $\frac{\mathbf{{K}^{T}} \cdot \mathbf{Q}}{\sqrt{{d}^{\prime}}}$ to obtain the mutual attention between nodes and update it by combining the normalized adjacency matrix $\mathbf{\bar{A}^r}$. Finally, we  perform a weighted sum of all the value matrices to aggregate the target node representations $\mathbf{Z}$. }
\label{fig4}
\end{figure*}

Each hyperedge type $ r \in \mathcal{T}_e$ in the heterogeneous hypergraph $\mathcal{G}_h$ defines a distinct semantic view. The corresponding heterogeneous hypergraph $\mathcal{G}^r_h$ within a single view can be seen as a view-specific subgraph of $\mathcal{G}_h$ sampled by $ \mathcal{E}^r_h $. To facilitate subsequent operations, we first project the features of target nodes $ \mathbf{X} \in \mathbb{R}^{|\mathcal{V}|\times d} $ into  $\mathbf{X}^{\prime} \in \mathbb{R}^{ |\mathcal{V}|\times d^{\prime}} $  via a linear transformation:
\begin{equation}
    \mathbf{X}^{\prime} = \textrm{MLP}(\mathbf{X}).
\end{equation}

To capture long-range node-node interactions within a single view, we then devise a node-level attention mechanism to learn effective node representations, inspired by the Transformer~\cite{r30}. Specifically, we use three learnable weight matrices $ \mathbf{W}^{Q}, \mathbf{W}^{K}, \mathbf{W}^{V} \in{ \mathbb{R}}^{{d}^{\prime} \times {d}^{\prime}}$, and multiply them with $\mathbf{X}^{\prime} $ to obtain the corresponding query $(\mathbf{Q})$, key $(\mathbf{K})$, and value $(\mathbf{V})$ matrices, where $ \mathbf{Q ,K,V}\in{ \mathbb{R}}^{{d}^{\prime}\times  {d}^{\prime}}$.
By computing $\mathbf{K}^{\mathrm{T}}\mathbf{Q} $, we obtain the similarity matrix $\bm{\alpha}^r$ between nodes, which is calculated as follows:
\begin{align}
 \mathbf{Q}=\mathbf{W}^{Q} \mathbf{X}^{\prime},  \hspace{0.7cm}  \mathbf{K}= &\mathbf{W}^{K} \mathbf{X}^{\prime}, \hspace{0.7cm}  \mathbf{V}=\mathbf{W}^{V} \mathbf{X}^{\prime} \\ 
    \bm{\alpha}^r &= \frac{\mathbf{K}^{\mathrm{T}} \mathbf{Q} }{\sqrt{{d}^{\prime}}},
\end{align}
where $\bm{\alpha}^r$ indicates node-level attention (i.e., the similarity between the target node and the other $n-1$ nodes). Then, the normalized adjacency matrix $\mathbf{\bar{A}}^r$ from one view is used to adjust the attention coefficients, which are further normalized using the softmax function: 
\begin{align}
\hat{\bm{\alpha} }^r &\! = \! \mathrm{Softmax}\left(  \! \bm{\alpha}^r \odot \mathbf{\bar{A}}^r \right),
\end{align}
where $\odot$ denoted the Hadamard product. The target node representations $\mathbf{Z}^r$ in a single view  can be then computed by:
\begin{equation}
\mathbf{Z}^r = {\hat {\bm{\alpha }}^r}  \mathbf{V}.
\end{equation}

To stabilize the learning process and better explore the potential relations between nodes, we extend the attention mechanism to multiple heads. The process is illustrated in Fig.~\ref{fig4}. Firstly, the node features $\mathbf{X}^{\prime} \in \mathbb{R}^{|\mathcal{V}| \times {d}^{\prime}} $ are scaled into $h$ sub-vector matrices $\mathbf{X}^{\prime (1)},\mathbf{X}^{\prime (2)}, \cdots ,\mathbf{X}^{\prime(h)} \in \mathbb{R}^{|\mathcal{V}| \times h_{d}}$, where $ {h}_{d}={d}^{\prime} / h $ denotes the number of attention heads. Similarly, for each attention head $h$, we also set three learnable weight matrices $ \mathbf{W}^{{Q},(h)}, \mathbf{W}^{K,(h)}, \mathbf{W}^{V,(h)}\in \mathbb{R}^{{h}_{d} \times {h}_{d} } $, and use attention mechanisms to update the information for each head.
Accordingly, the target node representations $\mathbf{Z}^r $ in a single view can be computed as:
\begin{equation}
\mathbf{Z}^r =\textrm{MLP}\left ( \textrm{cat}\left ( \mathbf{Z}^{r,(1)} ,\mathbf{Z}^{r,(2)},\cdots ,\mathbf{Z}^{r,(h)}\right ) \oplus \mathbf{X} \right ),
\end{equation}
where cat denotes the concatenate operation, $\oplus$ denotes the residual connection, and $\mathbf{X}$ denotes the original node feature matrix. The resulting node representations $ \mathbf{Z}^r$ learned within a single view are expected to capture high-order relations characterized by hyperedge type $r$. 

\subsection{Hyperedge-Level Attention }
\begin{figure}[htbp]
\centering
\includegraphics[width=0.45\textwidth]{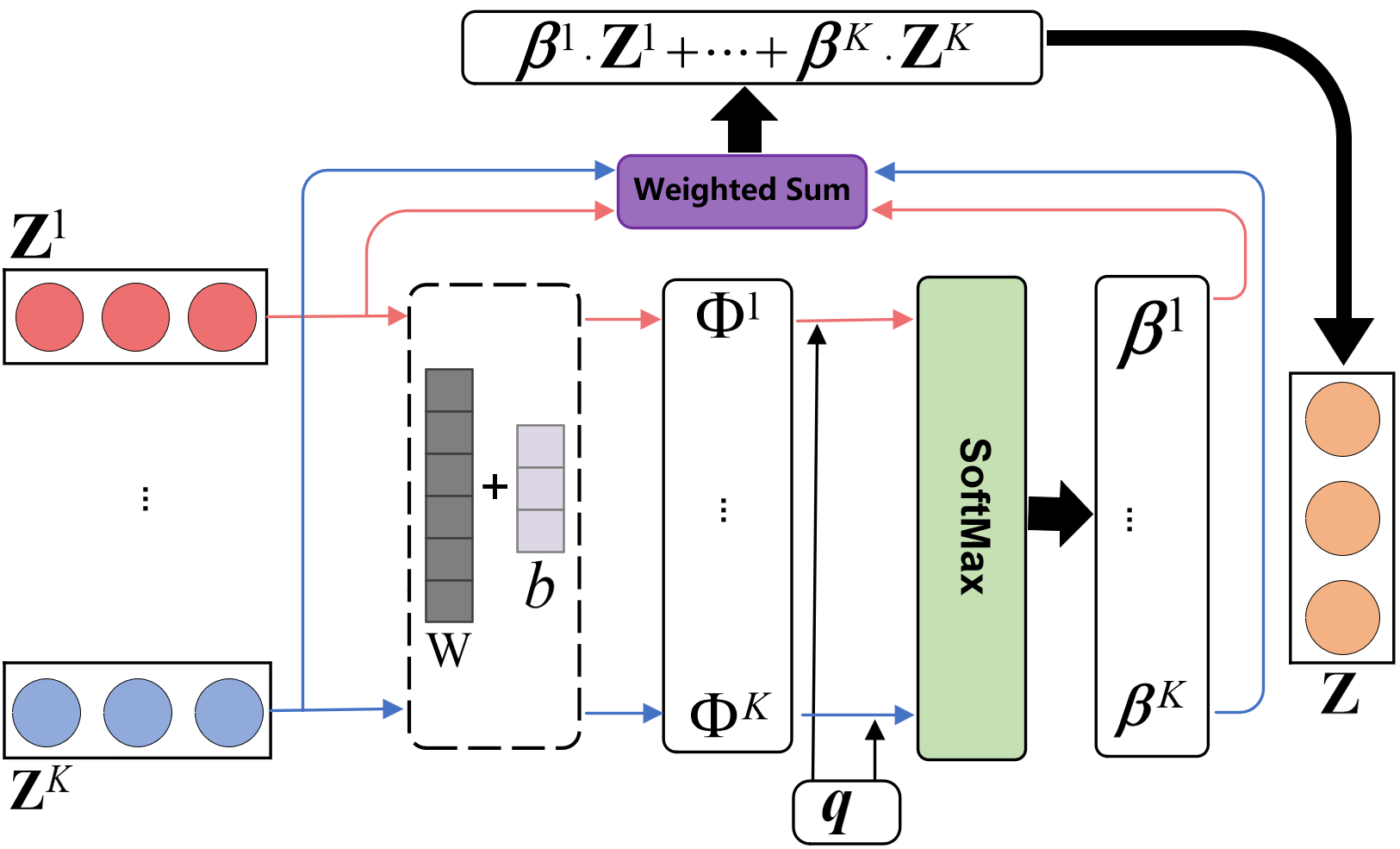}
\caption{Hyperedge-level attention. The weights of each hyperedge type are first learned through hyperedge-level attention, which are further normalized by the softmax function. The weighted sum of the semantic-specific node representations are obtained to generate the final node representations. }
\label{fg5}
\end{figure}

To capture rich and diverse semantics between target nodes across different hyperedge types, we propose a hyperedge-level attention mechanism that aggregates node representations learned from multiple views, each specified by one hyperedge type. 
Specifically, we aggregate node representations $\left \{ \mathbf{Z}^r\right \}_{r=1}^K$ across different views to produce the final target node representations $\mathbf{Z} $. The aggregation process is illustrated in Fig.~\ref{fg5}.

First, we apply a nonlinear transformation to node representations learned from each view and obtain the average representation of all target nodes in that view:
\begin{equation}
{\Phi }^r =\frac{1}{\left | \mathcal{V} \right |  } \sum \mathrm{tanh}\left ( \mathbf{W} \cdot \mathbf{Z}^r + \mathbf{b}\right ),
\end{equation}
where $\left | \mathcal{V} \right |$ is the number of target nodes,  $\mathbf{W}$ and $ \mathbf{b}$ are learnable parameters.  It is important to note that all the above parameters are shared among multiple views.

To quantify the importance of different hyperedge types, we introduce a learnable semantic-level attention vector $\mathbf{q}$, which is randomly initialized and optimized during training. After obtaining the importance of each hyperedge type, we normalize them using the softmax function. The weight of the hyperedge type ${r}$ is defined as $\beta^r$, given by:
\begin{align} 
\beta^r & = \frac{\mathrm{exp}\left (\mathbf{ q  }^\mathrm{T} \cdot {\Phi }^r \right )}{\displaystyle\sum_{r=1}^{K}\mathrm{exp}\left ( \mathbf{q}^{\mathrm{T}} \cdot {\Phi }^r  \right )}.
\end{align}

Finally, the learned weights are used as coefficients to fuse node representations learned from different views. The final node representations $\mathbf{Z}$ are obtained as follows:
\begin{equation}
\mathbf{Z}= \displaystyle\sum_{r=1}^{K} \beta^r \mathbf{Z}^r
\end{equation}

\subsection{Complexity Analysis}
Lastly, we analyze the time complexity of the proposed MGA-HHN model. The heterogeneous hypergraph construction based on symmetric meta-paths has a time complexity of $ \Large{O}\left ( KN\right )$, where $N$ denotes the number of target nodes and $K$ denotes the number of hyperedge types. The most computationally expensive operation is the calculation of node-level attention between nodes within each hypergraph, which requires a time complexity of $ \Large{O}( KN^2d\widetilde{d})$, where $d$ and $\widetilde{d}$ denote the dimensions of the input and output features, respectively. For hyperedge-level attention, computing the weights for $K$ types of hyperedges and performing normalization have a time complexity of $\Large{O}( KN\widetilde{d})$. Therefore, the overall time complexity of MGA-HHN is 
${\Large{O}}(KN)+\Large{O}( KN^2d\widetilde{d})+{\Large{O}} (KN\widetilde{d})\approx {\Large{O}} (KN^2d\widetilde{d})$.

\section{Experiments}
In this section, we evaluate the effectiveness of our proposed MGA-HHN model. We first describe the experimental setup, and then present experimental results and analysis on real-world  datasets. 
 
\subsection{Experimental Setup}
\subsubsection{Datasets}
We perform experiments on three real-world datasets from different domains. The statistics of these datasets are provided in Table~\ref{tab2}, where ``*" indicates the target node type in the task. 
\begin{itemize}
\item DBLP\footnote{\url{https://dblp.uni-trier.de.}} is a citation network in computer science, containing four types of nodes: authors, articles, conferences and terms. It also includes various types of relations, such as author-paper, paper-conference.

\item IMDB\footnote{\url{https://dl.acm.org.}} is a film rating network that records users’ preferences for different movies. It contains three different types of nodes: movies, actors and directors. It includes various relations such as movie–director, movie–actor.

\item ACM\footnote{\url{https://www.imdb.com.}} is a citation network extracted from the ACM database, containing four types of nodes: authors, articles, conferences and subjects. There are many types of relations, such as author-paper, paper-conference.
\end{itemize}


\begin{table}[h]
\caption{Statistics of datasets }
\label{tab2}
\resizebox{\columnwidth}{!}{
\begin{tabular}{cllll}
\toprule
Dataset              & \#\,Nodes         & \#\,Features  & Meta-paths                                                                                  & \#\,Classes              \\ \midrule
\multirow{4}{*}{DBLP} & Authors* ($A$): 4,057       & \multirow{3}{*}{334}  & \multirow{4}{*}{\begin{tabular}[c]{@{}l@{}} $APA$\\ $APCPA$ \\ $APTPA$\end{tabular}} & \multirow{4}{*}{4} \\ 
                      & Papers ($P$): 14,328        &  &                                                                                          &                    \\
                      & Conferences ($C$): 20      &     &                                                                                          &                    \\
                      & Terms ($T$): 7,723         &    &                                                                                          &                    \\  \midrule
\multirow{3}{*}{IMDB} & Movies* ($M$): 4,278        & \multirow{3}{*}{3,066} & \multirow{3}{*}{\begin{tabular}[c]{@{}l@{}} $MAM$ \\ $MDM$ \end{tabular}}                  & \multirow{3}{*}{3} \\
                      & Directors ($D$): 2,081      &  &                                                                                          &                    \\
                      & Actors ($A$): 5,257        &  &                                                                                          &                    \\   \midrule
\multirow{3}{*}{ACM}  & Papers* ($P$): 4,025        & \multirow{3}{*}{1,903} & \multirow{3}{*}{\begin{tabular}[c]{@{}l@{}} $PAP$ \\ $PSP$ \end{tabular}}                  &  \multirow{3}{*}{3} \\
                      & Authors ($A$): 17,431    &     &                                                                                       &                    \\
                      & Subjects ($S$): 73         &    &                                                                                          &                    \\    \bottomrule
\end{tabular}
}
\end{table}

\subsubsection{Baselines} 

We compare our proposed MGA-HHN model with state-of-the-art models from four representative categories: (1) Homogeneous GNNs including GCN~\cite{gcn} and GAT~\cite{gat}; (2) Hetergeneous GNNs including HAN~\cite{HAN} and MAGNN~\cite{magnn}; (3) Homogeneous hypergraph embedding models like HGNN~\cite{hgnn}; and (4) Heterogeneous hypergraph embedding models including HWNN~\cite{hwnn} and HGTN~\cite{li2023hypergraph}. The details of these baseline methods are described as follows:

\begin{itemize}
\item \textbf{GCN}~\cite{gcn}: A homogeneous GNN that performs spectral convolutions on graph structures.
\item \textbf{GAT}~\cite{gat}: A homogeneous GNN that performs spatial graph convolution with an attention mechanism to weigh neighbor features.
\item \textbf{HAN}~\cite{HAN}: A heterogeneous attention network that transforms a heterogeneous graph into several homogeneous graphs via given meta-paths. It employs both node-level and semantic-level  attention to calculate the importance scores of meta-path based neighbors, and then fuses node representations learned from different meta-path based subgraphs.
\item \textbf{MAGNN}~\cite{magnn}: An Hetergeneous GNN that extends HAN by considering both the meta-path based neighborhoods and the nodes within the meta-path instances.
\item \textbf{HGNN}~\cite{hgnn}: A homogeneous hypergraph convolutional network that generalizes spectral convolutions to hypergraph structures.
\item \textbf{HWNN}~\cite{hwnn}: A heterogeneous hypergraph embedding model that transforms the Fourier of hypergraph convolution into wavelet convolution to better model heterogeneous hypergraph.
\item \textbf{HGTN}~\cite{li2023hypergraph}: A heterogeneous hypergraph embedding model that uses an attention mechanism to learn the weights of different types of hypergraphs and discover useful meta-paths. It then utilizes a multi-scale attention module to aggregate node representations from high-order neighbors.
\end{itemize} 

\subsubsection{Implementation Details}

We implement MGA-HHN based on Pytorch and train it with 100 iterations on all datasets. The Adam optimizer \cite{adam} is used with a learning rate of 0.001. To further evaluate the performance of our model in semi-supervised learning tasks, we vary the ratio of training sets from 20\%, 40\%, 60\% to 80\%. For model training, we use an early stopping mechanism with patience of 10, i.e., the training process is stopped when the validation loss does not decrease for 10 consecutive epochs. Each experiment was repeated five times with different random seeds, and the average of the five runs was taken as the final result. All models are trained on the RTX A5000 (24GB) GPU.

\begin{table*}[t]
\caption{Performance comparison on the DBLP, IMDb and ACM datasets for the node classification task.}
\label{tab3}
\resizebox{\textwidth}{!}{
\renewcommand{\arraystretch}{1.1}
\begin{tabular}{l|l|c|cc|cc|ccc|c}
\toprule
Dataset                        & Metrics                   & Train & GCN   & GAT   & HAN   & MAGNN & HGNN  & HWNN  & HGTN  & MGA-HHN         \\  \midrule
\multirow{8}{*}{DBLP} & \multirow{4}{*}{Macro-F1} & 20\%  & 83.41 & 83.21 & 91.46 & 92.72 & 65.05 & 80.05 & 91.53 & \textbf{92.88} \\  
                               &                           & 40\%  & 86.30 & 84.99 & 92.07 & 93.19 & 70.47 & 82.06 & 92.06 & \textbf{93.31} \\
                               &                           & 60\%  & 87.55 & 85.26 & 92.47 & 93.37 & 74.29 & 83.20 & 92.56 & \textbf{93.56} \\ 
                               &                           & 80\%  & 88.39 & 85.48 & 92.43 & 93.67 & 72.74 & 84.29 & 93.05 & \textbf{94.43} \\  \cline{2-11}
                               & \multirow{4}{*}{Micro-F1} & 20\%  & 82.68 & 82.43 & 92.43 & 93.29 & 64.53 & 80.51 & 92.22 & \textbf{93.32} \\  
                               &                           & 40\%  & 85.72 & 84.33 & 92.70 & 93.68 & 70.22 & 82.78 & 92.73 & \textbf{93.70} \\ 
                               &                           & 60\%  & 86.69 & 84.59 & 92.77 & 93.89 & 74.11 & 83.94 & 93.21 & \textbf{93.89} \\ 
                               &                           & 80\%  & 87.89 & 84.82 & 92.86 & 94.25 & 72.44 & 85.07 & 93.47 & \textbf{94.62} \\  \hline
\multirow{8}{*}{IMDB}          & \multirow{4}{*}{Macro-F1} & 20\%  & 60.59 & 58.30 & 60.77 & 59.10 & 51.99 & 55.62 & 54.21 & \textbf{61.27} \\  
                               &                           & 40\%  & 63.30 & 61.97 & 61.29 & 60.55 & 59.00 & 58.34 & 55.64 & \textbf{63.41} \\ 
                               &                           & 60\%  & 65.33 & 63.65 & 64.17 & 60.85 & 62.79 & 59.73 & 57.08 & \textbf{65.85} \\ 
                               &                           & 80\%  & 66.95 & 65.54 & 66.49 & 62.30 & 64.23 & 63.44 & 59.40 & \textbf{68.46} \\  \cline{2-11} 
                               & \multirow{4}{*}{Micro-F1} & 20\%  & 59.56 & 58.66 & 61.17 & 59.01 & 51.64 & 56.21 & 62.74 & \textbf{61.75} \\ 
                               &                           & 40\%  & 62.07 & 60.87 & 62.41 & 60.64 & 58.99 & 59.31 & 63.06 & \textbf{63.30} \\ 
                               &                           & 60\%  & 64.39 & 62.41 & 64.76 & 60.90 & 62.84 & 60.04 & 63.83 & \textbf{65.83} \\ 
                               &                           & 80\%  & 66.15 & 64.39 & 65.43 & 62.46 & 64.42 & 63.76 & 65.75 & \textbf{68.48} \\  \hline
\multirow{8}{*}{ACM}           & \multirow{4}{*}{Macro-F1} & 20\%  & 91.30 & 91.03 & 91.52 & 90.07 & 71.33 & 71.38 & 90.70 & \textbf{91.51} \\ 
                               &                           & 40\%  & 92.12 & 91.35 & 91.78 & 91.42 & 74.56 & 72.88 & 91.63 & \textbf{92.60} \\ 
                               &                           & 60\%  & 92.25 & 92.25 & 92.43 & 91.71 & 78.83 & 73.14 & 92.18 & \textbf{93.09} \\ 
                               &                           & 80\%  & 93.39 & 93.39 & 92.44 & 92.11 & 80.50 & 73.50 & 93.49 & \textbf{93.64} \\  \cline{2-11} 
                               & \multirow{4}{*}{Micro-F1} & 20\%  & 91.37 & 91.05 & 91.48 & 90.01 & 71.53 & 76.24 & 90.65 & \textbf{91.54} \\ 
                               &                           & 40\%  & 92.19 & 91.66 & 91.65 & 91.39 & 73.57 & 76.75 & 91.68 & \textbf{92.59} \\ 
                               &                           & 60\%  & 92.32 & 92.31 & 92.21 & 91.68 & 78.85 & 76.81 & 92.03 & \textbf{93.11} \\ 
                               &                           & 80\%  & 93.74 & 93.44 & 92.29 & 92.08 & 80.11 & 77.52 & 93.29 & \textbf{93.81} \\  \bottomrule
\end{tabular}
}
\end{table*}

\subsection{Experiment Results and Analysis}
We first demonstrate the effectiveness of the proposed MGA-HHN model by comparing it with state-of-the-art baselines on various downstream tasks including node classification, node clustering, and visualization.

\subsubsection{Node Classification}
We first validate the performance of MGA-HHN on 
node classification using Micro-F1 and Macro-F1 scores as evaluation metrics. Table~\ref{tab3} presents a performance comparison of MGA-HHN and other baselines on the three datasets. 
From the results, we can see that MGA-HHN consistently outperforms all baselines across three datasets. For instance, compared to GCN, MGA-HHN achieves performance gains of 8.5\%, 1.9\%, and 0.5\%, on DBLP, IMDB, and ACM, respectively.  These improvements highlight the limitations of Homogeneous GNNs, which primarily focus on structure information while neglecting rich semantic information present in heterogeneous graphs.

In contrast to the homogeneous hypergraph model HGNN, MGA-HHN exhibits an average performance improvement of approximately 10\%. This is because HGNN constructs a hypergraph using a KNN-based approach, which relies only on node attribute similarity but ignores the structure information. 

Compared with heterogeneous GNNs such as HAN and MAGNN, MGA-HHN also demonstrates consistent performance improvements. On DBLP, IMDB, and ACM, MGA-HHN outperforms HAN by 1.4\%, 2.3\%, and 0.8\%, and MAGNN by 0.2\%, 1.3\%, and 1.5\%, respectively. These gains demonstrate the expressive power of meta-path based heterogeneous hypergraphs, which can capture diverse high-order relations guided by task-relevant meta-paths.

Finally, in comparison with heterogeneous hypergraph models, HWNN and HGTN, MGA-HHN achieves the average relative improvements of 15.7\% and 2.6\%, respectively. The significance margin over HWNN can be attributed to two key factors: a) MGA-HHN represents heterogeneous nodes as hyperedges, thereby better preserving the semantic context from the original heterogeneous graph; and b) MGA-HHN utilizes a transformer-based node-level attention to mechanism that enables each node to attend to every other node, rather than being limited to local neighborhoods, as in many other methods. 
\begin{figure*}[t]
  \centering
  \begin{subfigure}[b]{0.45\textwidth}
    \includegraphics[width=\textwidth]{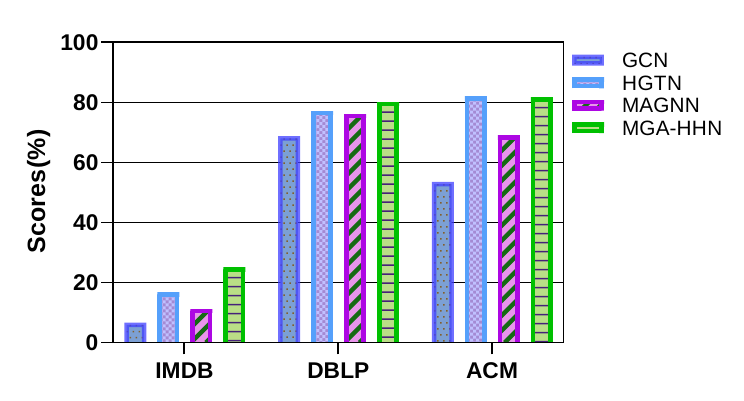} 
    \caption{ARI}
    \end{subfigure}
    \hfill
  \begin{subfigure}[b]{0.45\textwidth}
    \includegraphics[width=\textwidth]{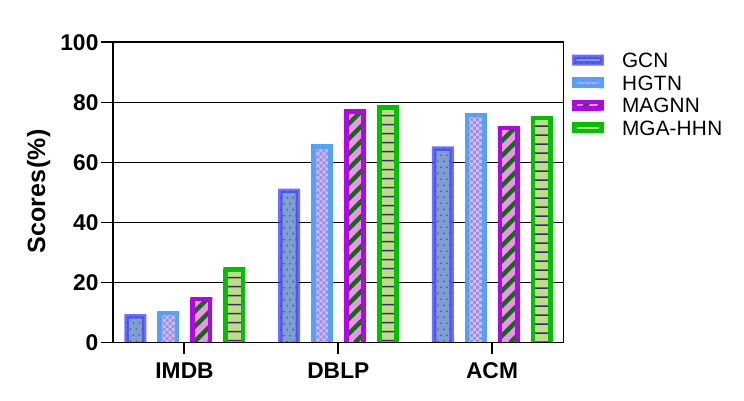}
    \caption{NMI}
    \end{subfigure}
   \caption{Performance comparison of node clustering results.}
   \label{6}
\end{figure*}

\begin{figure*}[t]
  \centering
  \begin{subfigure}[b]{0.23\textwidth}
    \includegraphics[width=\textwidth]{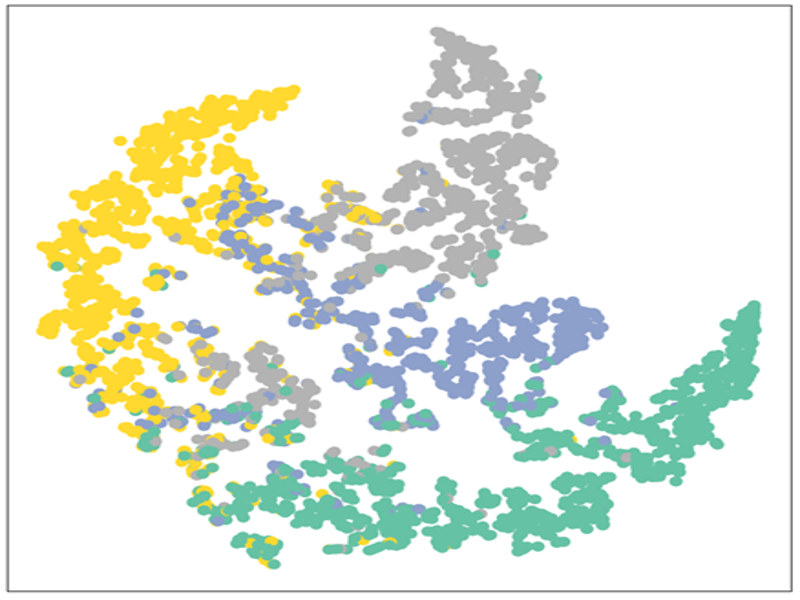}
    \caption{GCN}
  \end{subfigure}
  \hfill
  \begin{subfigure}[b]{0.23\textwidth}
    \includegraphics[width=\textwidth]{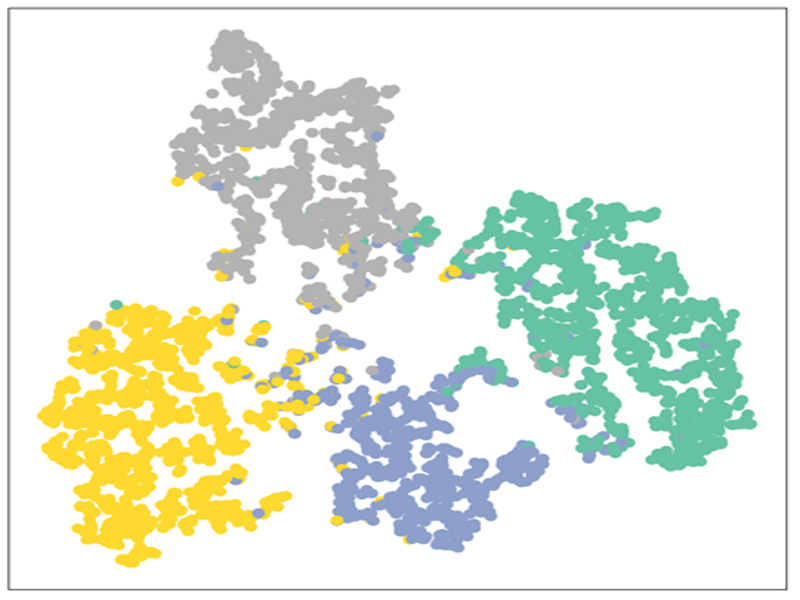}
    \caption{MAGNN}
  \end{subfigure}
  \hfill
  \begin{subfigure}[b]{0.23\textwidth}
    \includegraphics[width=\textwidth]{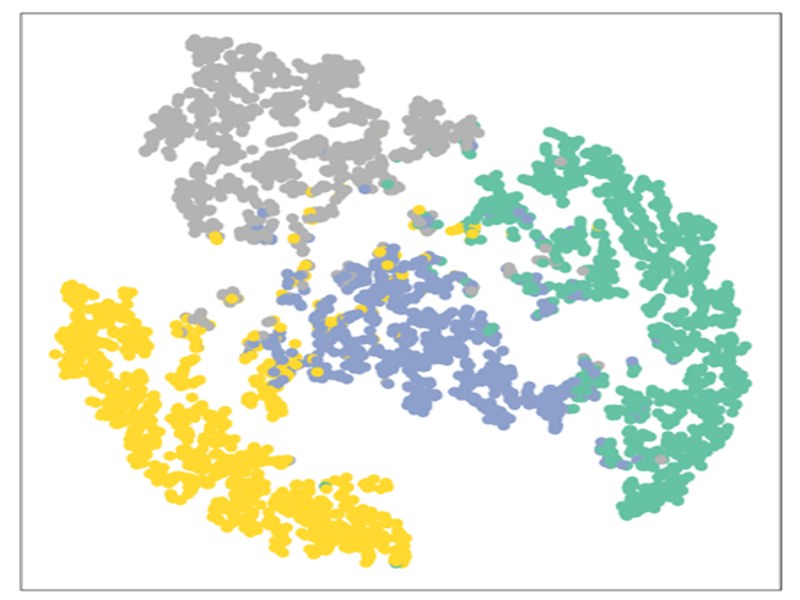}
    \caption{HGTN}
  \end{subfigure}
  \hfill
  \begin{subfigure}[b]{0.23\textwidth}
    \includegraphics[width=\textwidth]{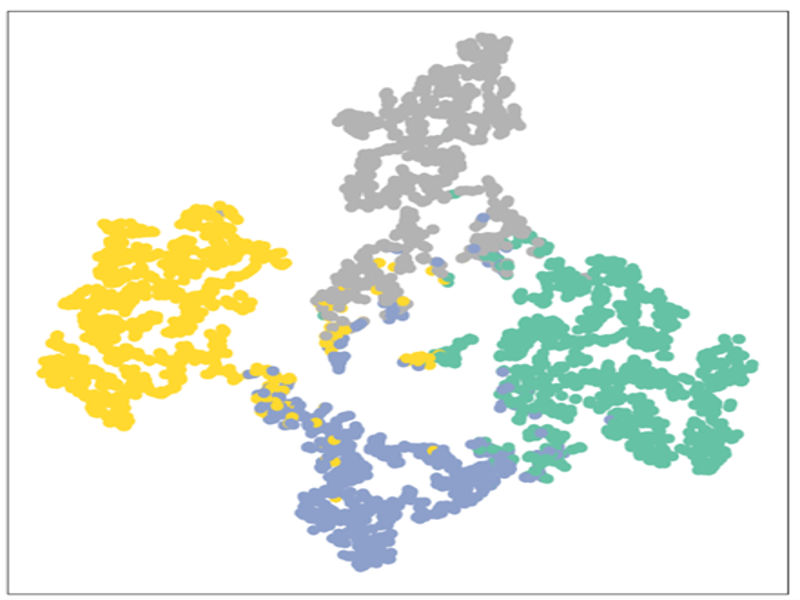}
    \caption{MGA-HHN}
    \end{subfigure}
  \caption{Visualization of latent node representations generated by different models on DBLP.}
  \label{fg7}
\end{figure*}

\subsubsection{Node Clustering}
We then evaluate the quality of the learned node representations through node clustering.

We apply k-means clustering to the target node representations learned by MGA-HHN, HGTN, MAGNN, and GCN on DBLP, IMDB and ACM datasets, respectively. The number of clusters is set to the number of target node class labels on each dataset. 
To assess the clusteirng performance, we use two widely adopted metrics: Normalized Mutual Information (NMI) and Adjusted Rand Index (ARI). 
NMI measures the mutual information between the predicted clustering and ground-truth labels, while ARI measures the consistency between the predicted clustering and ground-truth clusters. These two metrics provide a comprehensive assessment of the clustering quality. 

The experimental results are shown in Fig.~\ref{6}.
Among the baselines, HGTN generally performs the best on the three datasets. In contrast, MGA-HHN significantly outperforms HGTN. For example, on DBLP, MGA-HHN improves the NMI and ARI scores by 19.6\% and 18.3\%, respectively. These results indicate that MGA-HHN is capable of capturing rich structural and semantic information, leading to more informative and fine-grained node representations.

\subsubsection{Visualization}
To provide a more intuitive comparison between MGA-HHN and other methods, we also visualize the quality of the latent node representations learned by different models by projecting them into a two-dimensional space using t-SNE. Specifically, we visualize the author representations generated by four models---GCN, MAGNN, HGTN and MGA-HHN---on the DBLP dataset, with different colors indicating distinct research areas. 

As shown in Fig.~\ref{fg7}, the homogeneous GCN performs poorly, as the representations of authors from different research areas are heavily mixed. Compared with MAGNN and HGTN, MGA-HHN exhibits a much clearer clustering of authors with the same research area and better separated boundaries between different research areas.

\subsection{Ablation Studies}

To further investigate the contribution of each component in our proposed MGA-HHN model, we perform ablation experiments on three datasets. In particular, we compare our full model with four variants as follows. The ablation results are shown in Table~\ref{tab4}.

\begin{table*}[htbp]
\caption{Ablation study on the node classification task.}
\label{tab4}
\resizebox{\textwidth}{!}{
\begin{tabular}{@{\hspace{1\tabcolsep}}lcccccc@{}}
\toprule
\multicolumn{1}{@{\hspace{1\tabcolsep}}l}{Datasets}        & \multicolumn{2}{c}{DBLP} & \multicolumn{2}{c}{ACM} & \multicolumn{2}{c}{IMDB} \\ \cmidrule(lr){1-1} \cmidrule(lr){2-3} \cmidrule(lr){4-5} \cmidrule(lr){6-7}
Metrics(\%)     & Micro-F1    & Macro-F1   & Micro-F1   & Macro-F1   & Micro-F1    & Macro-F1   \\ \midrule
w/o hypergraph  & 91.75       & 91.33      & 92.69      & 92.65      & 65.83       & 65.61      \\
w/o node-level attention    & 93.85       & 93.60      & 91.02      & 91.03      & 67.24       & 67.11      \\
w/o hyperedge-level attention   & 93.85       & 93.54      & 93.16      & 93.08      & 66.67       & 66.67      \\
MGA-HHN hyperedge-concate & 84.90       & 84.05      & 89.62      & 89.49      & 64.00       & 63.85      \\
MGA-HHN (full model)         & \textbf{94.65}       & \textbf{94.43}      & \textbf{93.55}      & \textbf{93.50}      & \textbf{68.48}       & \textbf{68.46}      \\ \bottomrule
\end{tabular}
}
\end{table*}

\begin{figure*}[t]
  \centering
  \begin{subfigure}[b]{0.31\textwidth}
    \includegraphics[width=\textwidth]{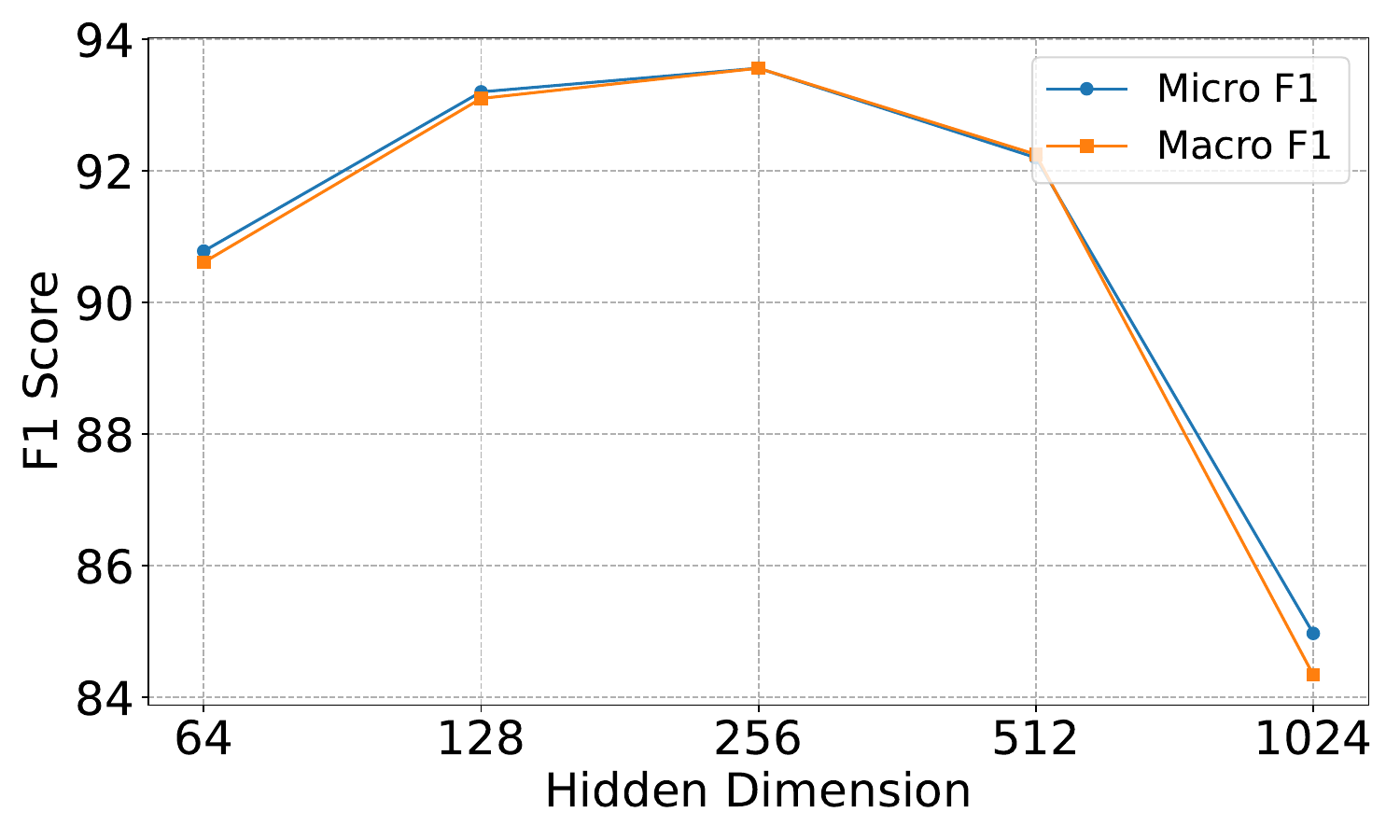}
    \caption{ACM}
    \end{subfigure}
  \hfill
  \begin{subfigure}[b]{0.31\textwidth}
    \includegraphics[width=\textwidth]{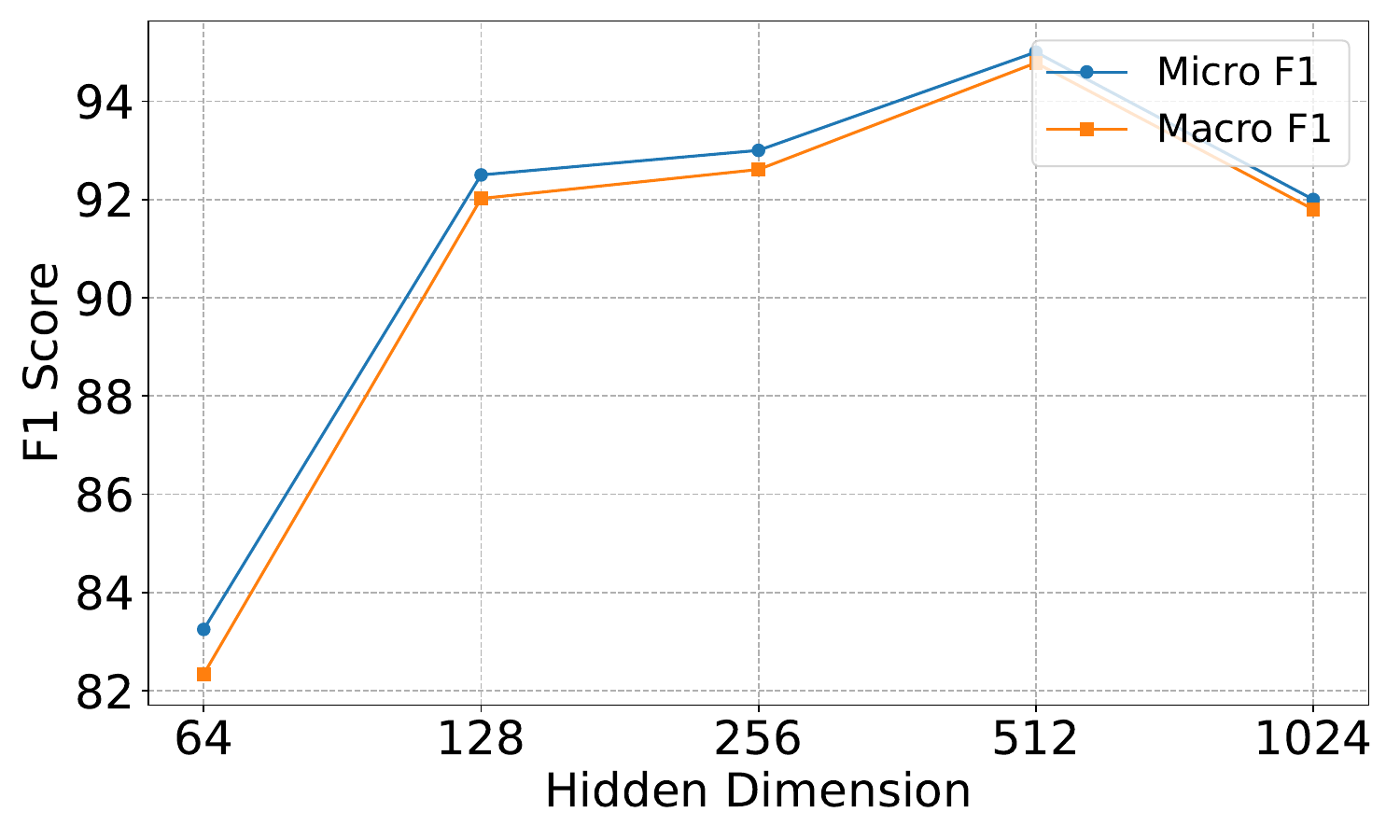}
    \caption{DBLP}
    \end{subfigure}
  \hfill
\begin{subfigure}[b]{0.31\textwidth}
    \includegraphics[width=\textwidth]{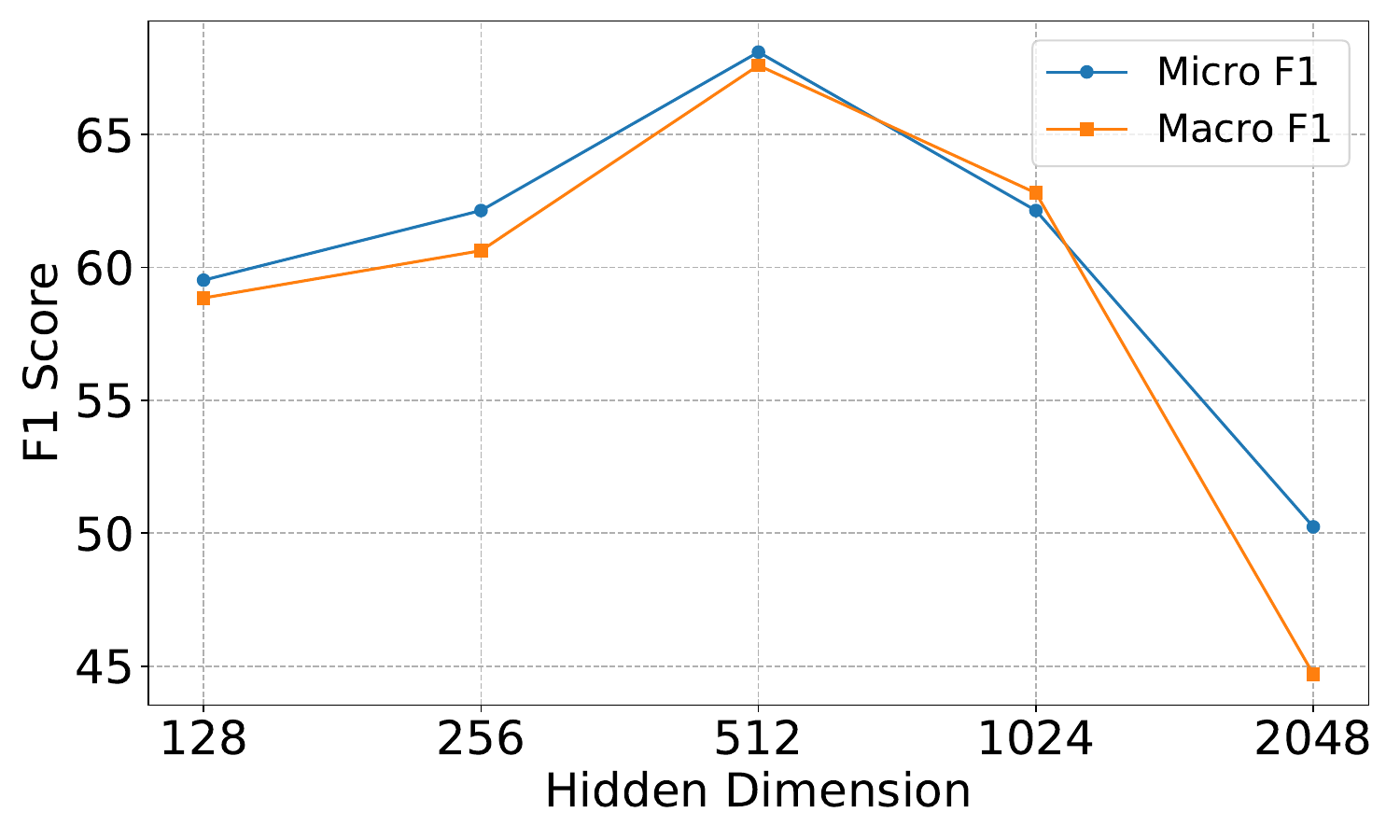}
    \caption{IMDB}
  \end{subfigure}
  \caption{Effect of hidden dimension on model performance. }
  \label{fg8}
\end{figure*}

\begin{figure*}[t]
  \centering
  \begin{subfigure}[b]{0.31\textwidth}    \includegraphics[width=\textwidth]{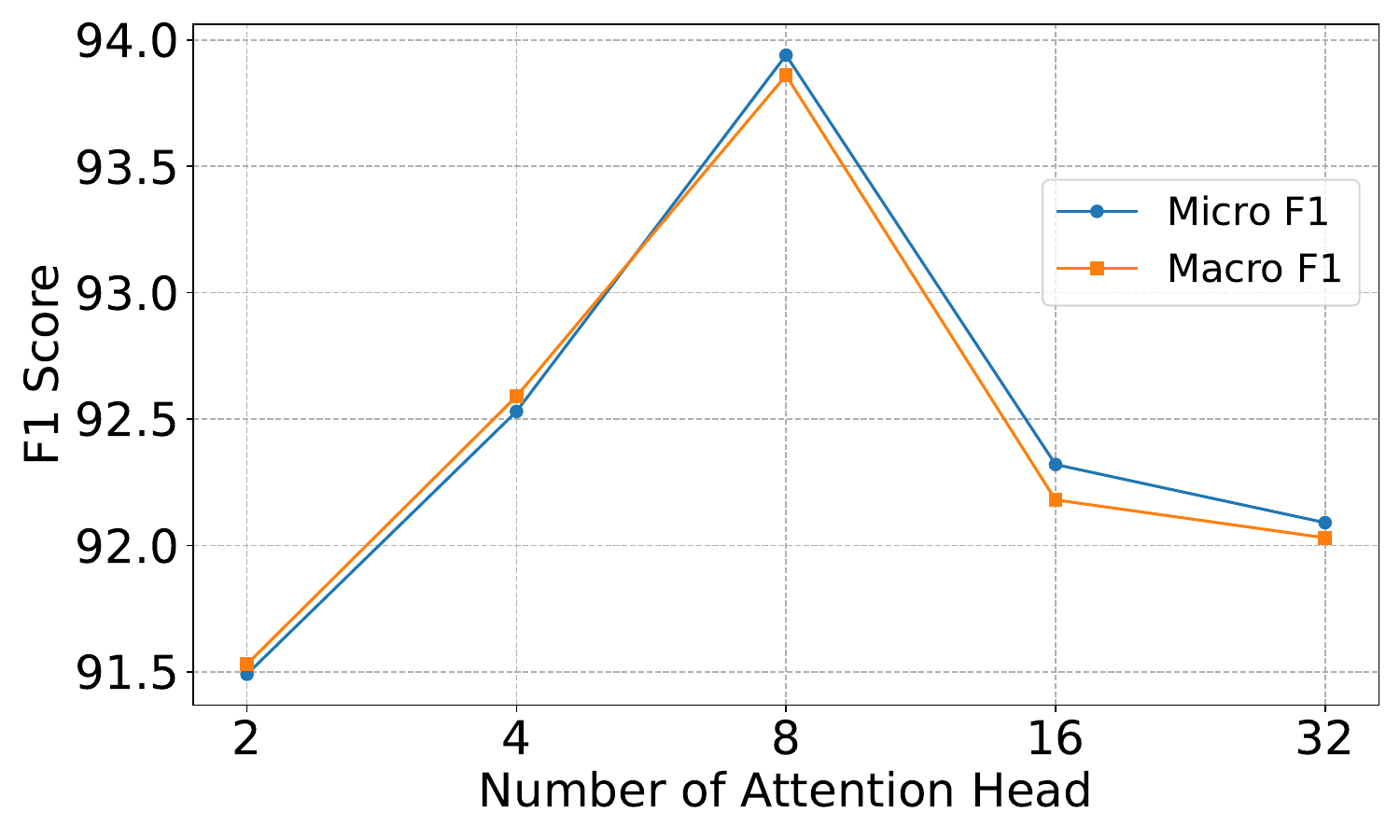}
    \caption{ACM}
    \end{subfigure}
  \hfill
  \begin{subfigure}[b]{0.31\textwidth}    \includegraphics[width=\textwidth]{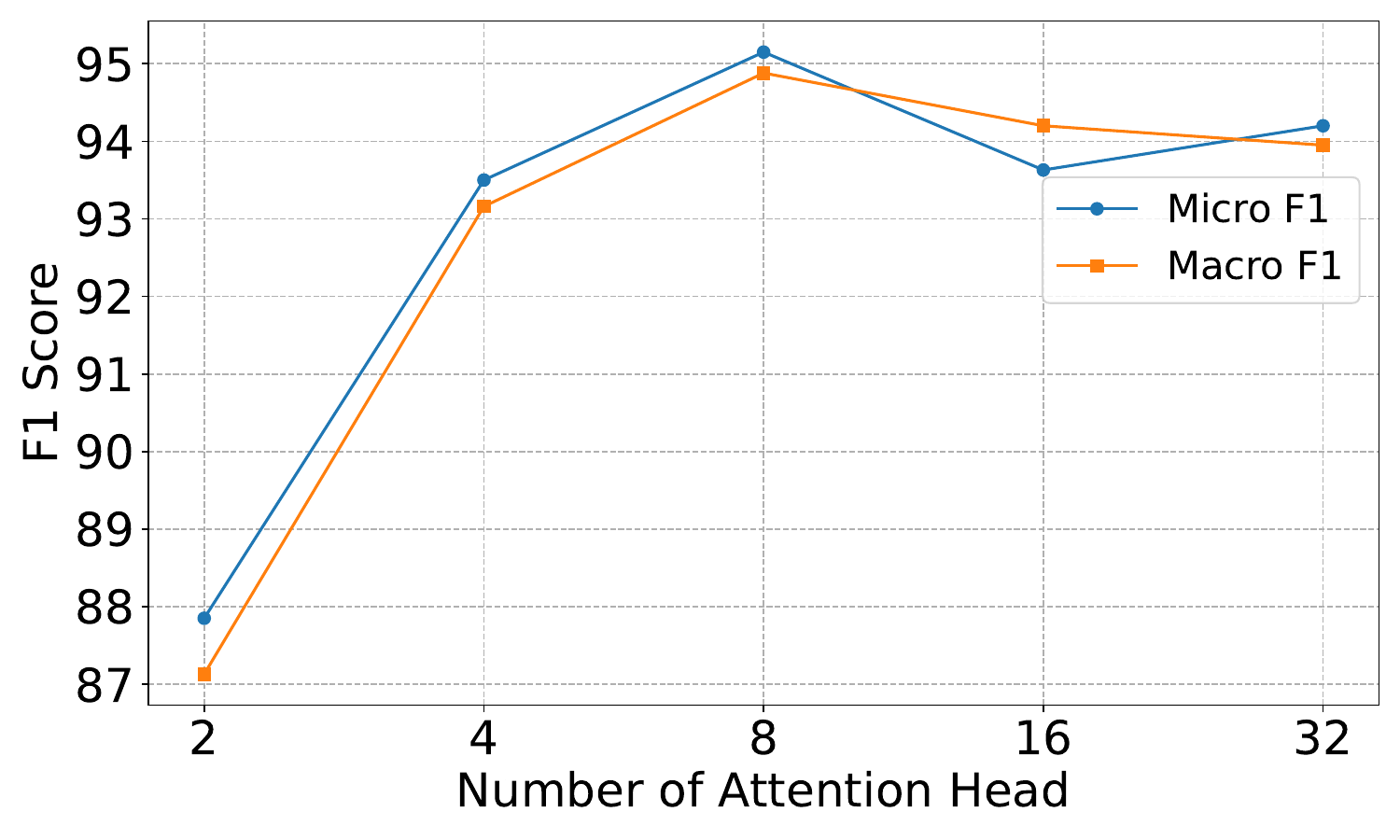}
    \caption{DBLP}
    \end{subfigure}
  \hfill
  \begin{subfigure}[b]{0.31\textwidth}    \includegraphics[width=\textwidth]{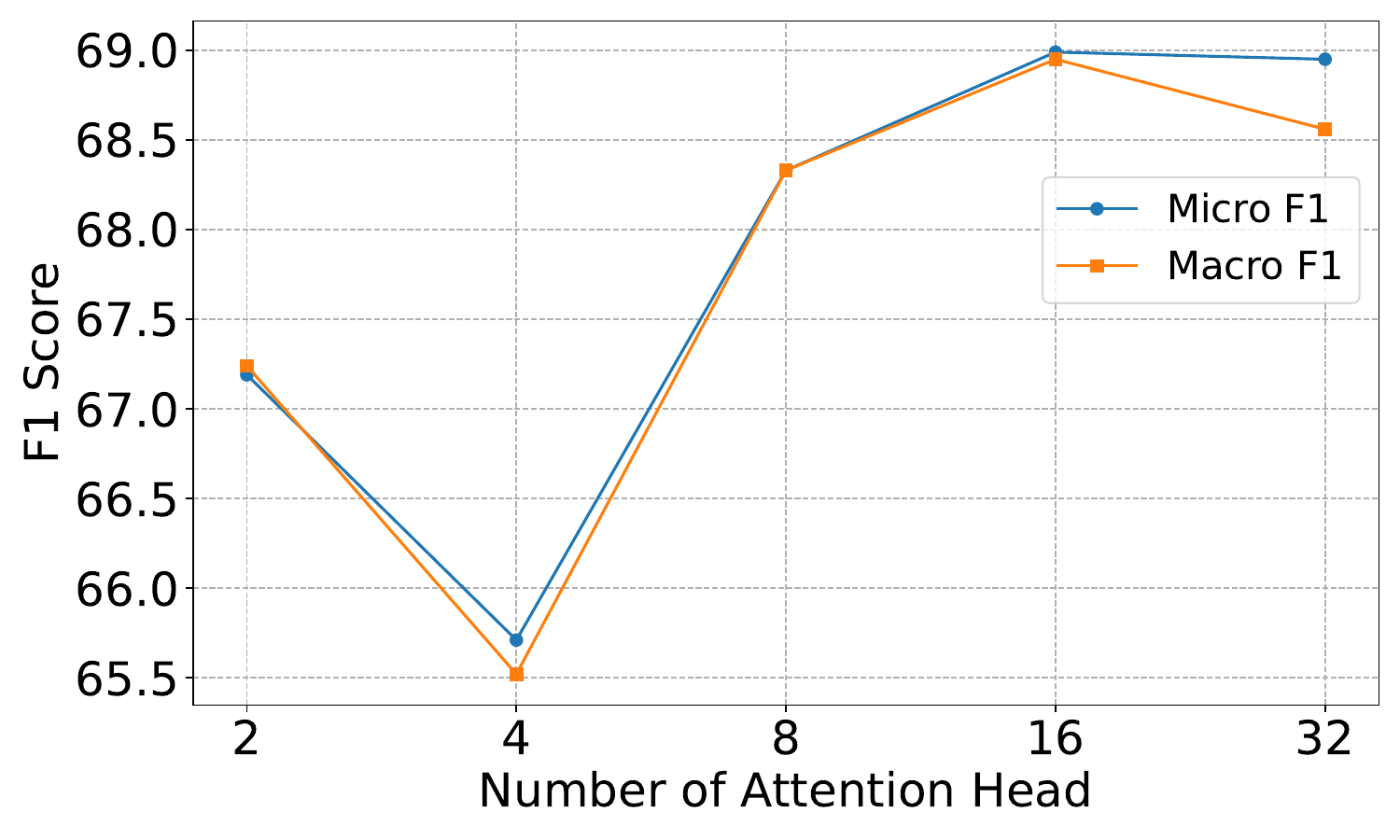}
    \caption{IMDB}
    \end{subfigure}
  \caption{Effect of attention head $h$ on model performance. }
  \label{fg9}
\end{figure*}

\textbf{w/o hypergraph} indicates the variant by replacing the hypergraph with a conventional graph. This ablation leads the F1 scores of the model to decrease by 3.3\%, 3.5\% and 1.3\% on DBLP, IMDB and ACM, respectively. This experiment indicates that constructing hypergraphs using meta-paths can explicitly preserve the high-order relations inherent in heterogeneous graphs.
    
\textbf{w/o Node-level Attention} refers to the variant that substitutes multi-head node-level attention with a hypergraph convolution in HGNN. On DBLP, IMDB and ACM, the removal of multi-head node-level attention results in a decline of F1 score by 0.9\%, 1.9\% and 2.7\%, respectively. This suggests that utilizing multi-head node-level attention can effectively capture long-range dependencies when the output of an MPNN depends on representations of distant nodes interacting with each other.

\textbf{w/o Hyperedge-level attention} indicates the variant in which hyperedge-level attention in the MGA-HHN model is replaced with where only one semantic (view) is considered, without taking into account multiple semantics or views. This ablation results in a pronounced drop in model performance. Specifically, the F1 score decreases markedly by 10.7\% on DBLP, and by approximately 6.6\% and 2.2\% on IMDB and ACM, respectively. These results highlight the important role of hyperedge-level attention in enabling our model to learn more expressive node representations that better account for the heterogeneity of abundant node/edge feature distributions.

\textbf{MGA-HHN Hyperedge-concate} refers to the variant of our model in which hyperedge-level attention is removed and replaced with a simple concatenation operation. The results show that this ablation leads to a performance decline in the F1 score by 0.8\%, 3.0\% and 0.6\% on DBLP, IMDB and ACM, respectively. These findings reveal that hyperedge-level attention can adaptively weigh the importance of different semantics. By putting more weights on informative semantics and weakening unimportant ones, hyperedge-level attention enables more effective heterogeneous graph representation learning.

\subsection{Hyper-Parameter Sensitivity Study}
We further conduct experiments to investigate the effect of two hyper-parameters---the dimension of the final representation and the number of attention heads---on the performance of MGA-HHN. We perform node classification tasks on the DBLP, ACM and IMDB datasets. 

\subsubsection{Effect of Hidden Dimension}
We examine the sensitivity of node classification performance with respect to different values of hidden dimensions across three datasets. The results shown in Fig.~\ref{fg8} show that the optimal hidden dimension varies across different datasets. 
In general, increasing the hidden dimension initially improves the performance. However, when beyond a certain point, further increases in hidden dimension lead to a performance drop. This can be attributed to the fact that MGA-HHN requires an appropriate hidden dimension to effectively encode semantic information in heterogeneous graphs. To be specific, MGA-HNN achieves the best node classification performance when the hidden dimension is set to 256, 512, and 512 on ACM, DBLP and IMDB, respectively. If the hidden dimension is too small, the model is unable to capture sufficient knowledge. Conversely, a too large hidden dimension may decentralize the model's focus on learning meaningful information.

\subsubsection{ Effect of Number of Attention Heads}
To study the impact of the number of attention heads in our proposed model, we perform experiments using 80\% nodes from the DBLP, IMDB and ACM datasets for training. We vary the number of attention heads from 2, $2^2$, $2^3$, $2^4$ to $2^5$, and report the results of node classification in Fig.~\ref{fg9}.

The results show that, on DBLP and ACM, the performance of our model improves as the number of attention heads $h$ increases, achieving the best performance when $h=8$. However, a further increase in $h$ leads to a decline in model performance. On IMDB, our model performs best when $h=16$. These results suggest that using multiple attention heads can augment node features, allowing messages from non-adjacent nodes to propagate across the network without significant distortion. 

\begin{figure*}[t]
  \centering
  \includegraphics[trim=0cm 6cm 1cm 7cm, clip,width=\textwidth]{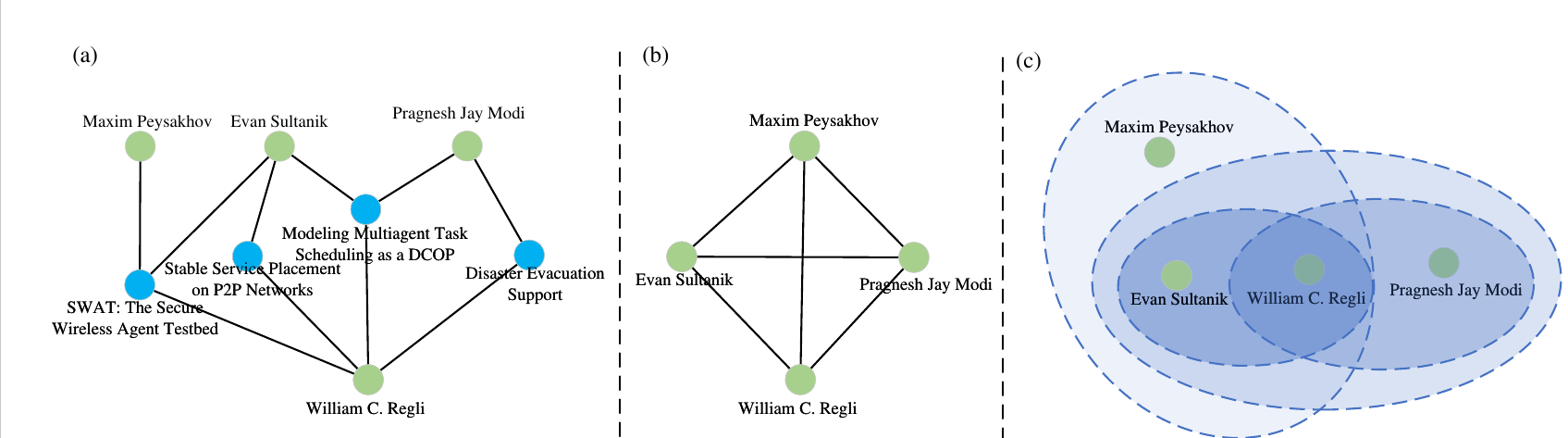}
  \caption{A case study based on the DBLP dataset. (a) A subgraph extracted as a heterogeneous graph. The green nodes represent authors, while the blue nodes represent papers. (b) Meta-path $APA$ based conventional graph. (c) Meta-path $APA$ based hypergraph.}
\label{10}
\end{figure*}

\begin{figure*}[t]
  \centering
  \begin{subfigure}[b]{0.45\textwidth}
  \includegraphics[width=\textwidth]{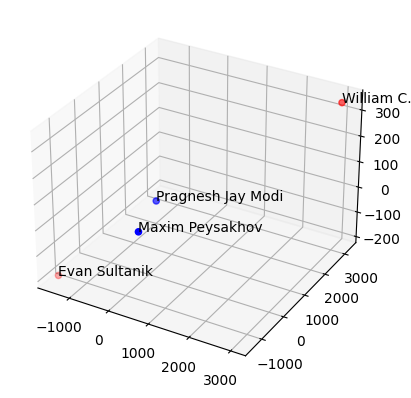} 
  \caption{HAN}  
  \end{subfigure}
  \begin{subfigure}[b]{0.45\textwidth}
  \includegraphics[width=\textwidth]{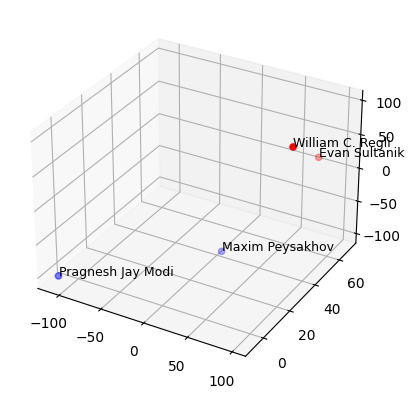}
  \caption{MGA-HHN}
  \end{subfigure}
\caption{t-SNE visualization of the learned representations of author nodes.}
\label{fig11}
\end{figure*}

\subsection{Case Study}

Lastly, to intuitively demonstrate the effectiveness of hypergraph structures constructed in MGA-HHN, we conduct a case study on the DBLP dataset to visually analyze the learned node representations. For simplicity, we extract a subgraph from DBLP that includes only two node types: Authors ($A$) and Papers ($P$), as shown in Fig.~\ref{10}(a). Based on a specific meta-path schema $APA$, we then respectively construct a conventional graph and a hypergraph from this subgraph, as illustrated in Fig.~\ref{10}(b) and Fig.~\ref{10}(c).  

To compare the expressive power of different graph structures, we apply HAN and MGA-HHN to learn the representations of all author nodes in the subgraph shown in Fig.~\ref{10}(a). HAN is a representative learning model built upon a conventional graph derived from the meta-path $APA$, while MGA-HHN learns node representations using a hypergraph constructed from the same meta-path. We then visualize the learned author representations using t-SNE, as shown in Fig.~\ref{fig11}. Compared to HAN, the representations learned by MGA-HHN exhibit a smaller distance between the author nodes \textit{William C. Regli} and \textit{Evan Sultanik}, indicating that these two author nodes are more similar. This is evident in the original subgraph, where the two authors co-author multiple papers. However, in the meta-path based conventional graph there is only one edge connecting these two author nodes. In contrast, the meta-path based hypergraph includes three hyperedges that contain both authors. This case study illustrates that using hypergraphs enables MGA-HHN to more effectively capture complex semantic relationships and thus learn richer and more informative node representations.

\section{conclusion}

In this paper, we proposed a novel method,  MGA-HHN, for learning node representations in heterogeneous graphs. MGA-HNN first constructs meta-path based heterogeneous hypergraphs to explicitly preserve the rich, high-order relations inherent in a heterogeneous graph. It then employs a multi-granular attention mechanism is devised at two levels: a transformer-based node-level attention module that learns long-range dependencies, ensuring that information from distant nodes is not excessively attenuated when propagating across the network; and a hyperedge-level attention module that adaptively fuses node representations from different semantic views into a unified representation. Extensive experiments on three heterogeneous graph datasets demonstrate that MGA-HHN outperforms baselines on node classification, node clustering and visualisation tasks. 

For future work, we plan to further explore the over-squashing phenomenon in heterogeneous graphs and develop theoretical insights that facilitate effective message passing from non-adjacent nodes without distortion. Additionally, we will investigate methods for automatically generating informative meta-paths to construct heterogeneous hyperedges more effectively.


\bibliographystyle{IEEEtran}
\bibliography{references}

\end{document}